\documentclass[10pt,twocolumn,letterpaper]{article}

\usepackage{cvpr}              

\usepackage[dvipsnames]{xcolor}

\usepackage{graphicx}
\usepackage{subcaption}
\usepackage{bm}
\usepackage{booktabs}
\usepackage{siunitx}
\usepackage{colortbl, color}

\usepackage{comment}
\usepackage[accsupp]{axessibility}

\definecolor{cvprblue}{rgb}{0.21,0.49,0.74}
\usepackage[pagebackref,breaklinks,colorlinks,citecolor=cvprblue]{hyperref}

\def\paperID{*****} 
\def\confName{CVPR}
\def\confYear{2024}

\begin{document}

\title{Deep Portrait Quality Assessment. A NTIRE 2024 Challenge Survey}

\author{
Nicolas Chahine~$^{*}$ \and
Marcos V. Conde~$^{*\dagger}$ \and
Daniela Carfora~$^{*}$ \and
Gabriel Pacianotto~$^{*}$ \and
Benoit Pochon~$^{*}$ \and
Sira Ferradans~$^{*}$ \and
Radu Timofte~$^{*}$ \and
Zhichao Duan \and
Xinrui Xu \and
Yipo Huang \and
Quan Yuan \and
Xiangfei Sheng \and
Zhichao Yang \and
Leida Li \and
Haotian Fan \and
Fangyuan Kong \and
Yifang Xu \and
Wei Sun \and
Weixia Zhang \and
Yanwei Jiang \and
Haoning Wu \and
Zicheng Zhang \and
Jun Jia \and
Yingjie Zhou \and
Zhongpeng Ji \and
Xiongkuo Min \and
Weisi Lin \and 
Guangtao Zhai \and
Xiaoqi Wang \and 
Junqi Liu \and 
Zixi Guo \and 
Yun Zhang \and
Zewen Chen \and
Wen Wang \and
Juan Wang \and
Bing Li
}

\twocolumn[{
\vspace{-3em}
\maketitle
\begin{center}
\setlength{\tabcolsep}{1pt}
\begin{tabular}{c c c c}
    \includegraphics[width=0.245\linewidth]{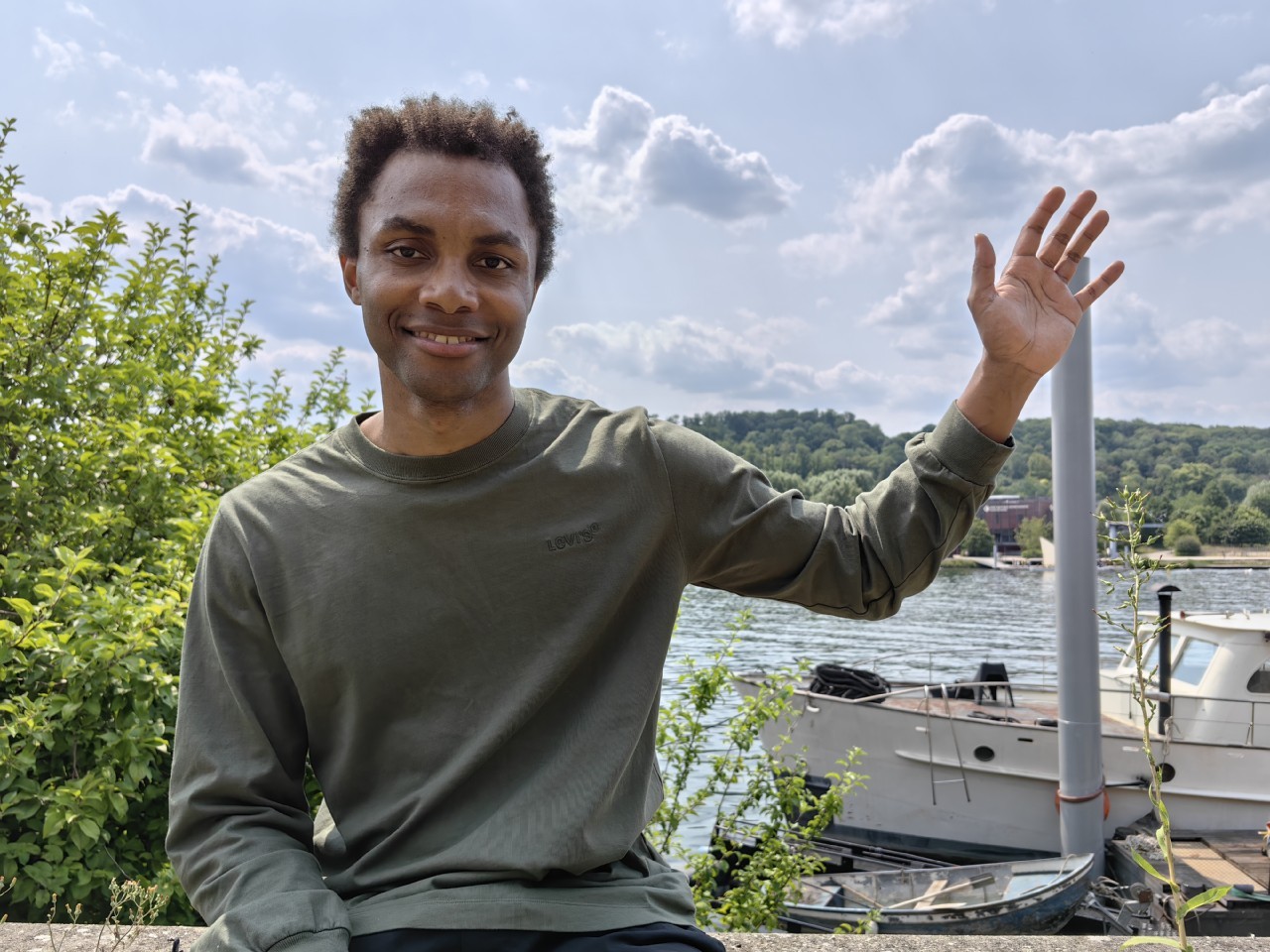} &
    \includegraphics[width=0.245\linewidth]{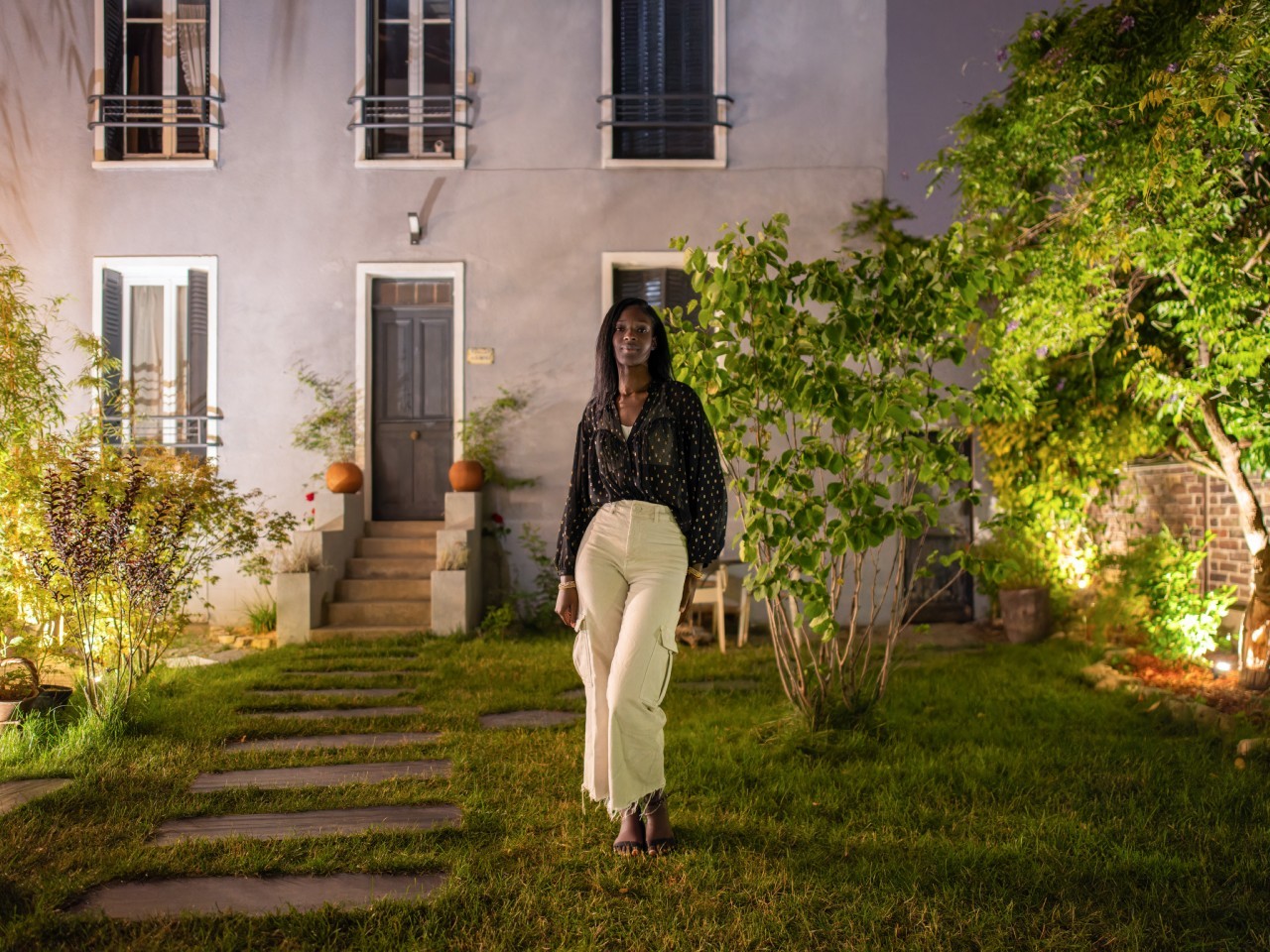} &
    \includegraphics[width=0.245\linewidth]{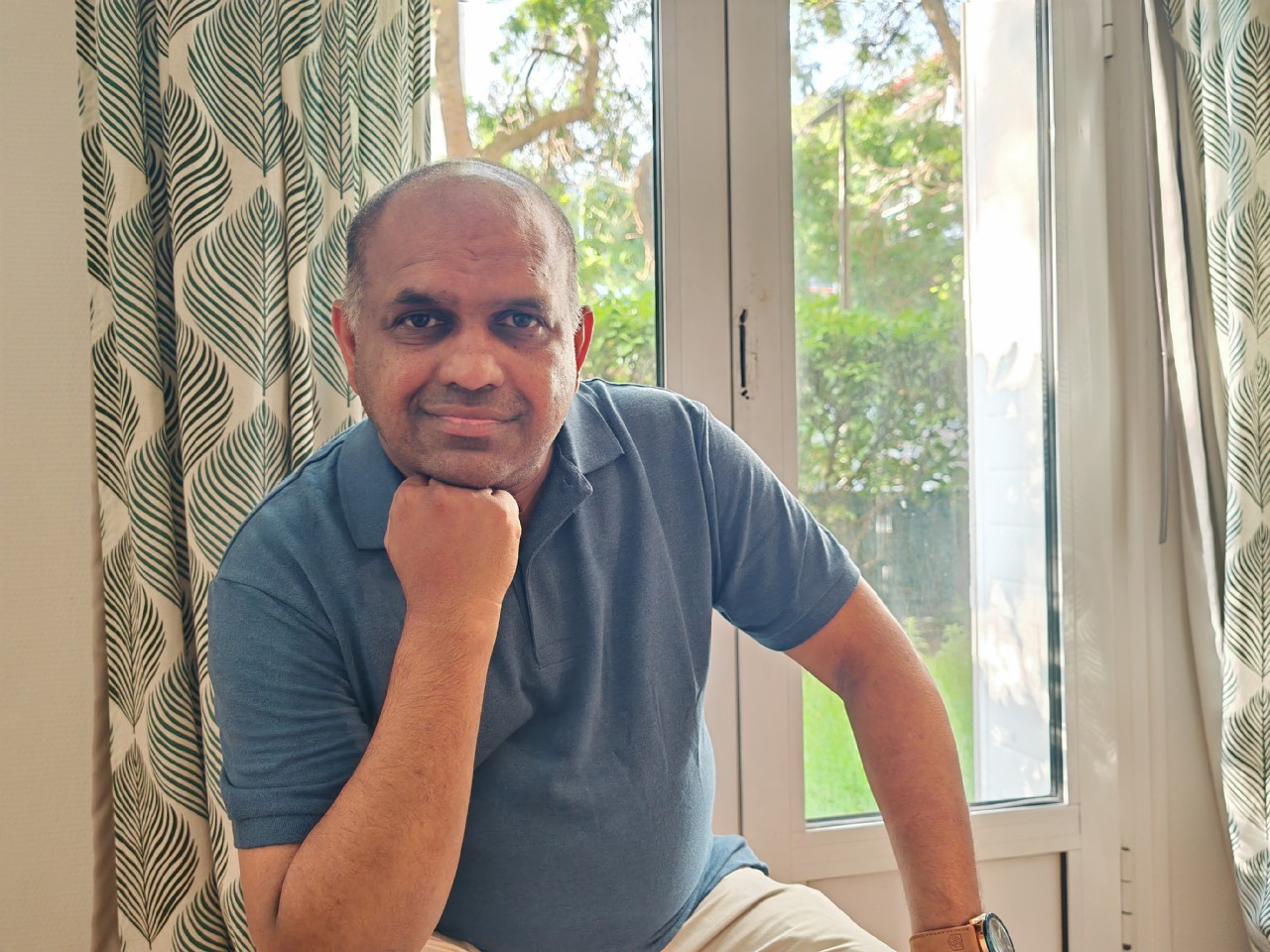} &
    \includegraphics[width=0.245\linewidth]{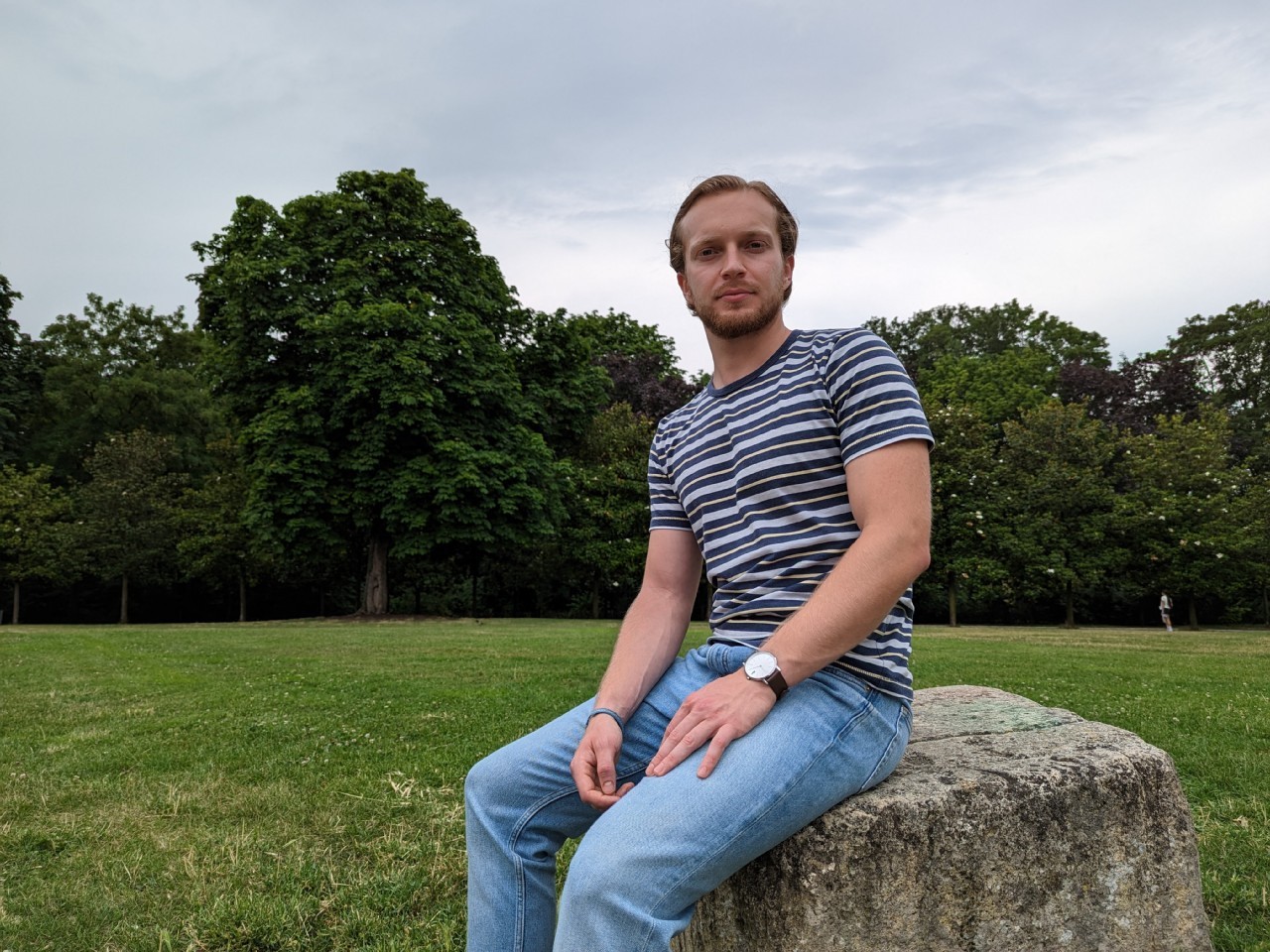} \\
\end{tabular}
\captionof{figure}{Sample portraits from the \emph{NTIRE 2024 Portrait Quality Assessment Challenge} testing set.}
\label{fig:samples}
\end{center}
\vspace{2em}
}]


\let\thefootnote\relax\footnotetext{$*$ Marcos V. Conde, Nicolas Chahine, Daniela Carfora, Gabriel Pacianotto, Sira Ferradans, Benoit Pochon, Radu Timofte are the challenge organizers, while the other authors participated in the challenge. \\

\noindent Marcos V. Conde ($\dagger$ corresponding author) and Radu Timofte are with University of W\"urzburg, CAIDAS \& IFI, Computer Vision Lab.\\

\noindent Nicolas Chahine, Daniela Carfora, Gabriel Pacianotto, Sira Ferradans, Benoit Pochon are with DXOMARK.\\

\noindent NTIRE 2024 webpage:~\url{https://cvlai.net/ntire/2024}\\
Code and Dataset:~\url{https://github.com/DXOMARK-Research/PIQ2023}
} 

\begin{abstract}
This paper reviews the NTIRE 2024 Portrait Quality Assessment Challenge, highlighting the proposed solutions and results. This challenge aims to obtain an efficient deep neural network capable of estimating the perceptual quality of real portrait photos. The methods must generalize to diverse scenes and diverse lighting conditions (indoor, outdoor, low-light), movement, blur, and other challenging conditions. In the challenge, 140 participants registered, and 35 submitted results during the challenge period. The performance of the top 5 submissions is reviewed and provided here as a gauge for the current state-of-the-art in Portrait Quality Assessment.
\end{abstract}

\setlength{\abovedisplayskip}{1pt}
\setlength{\belowdisplayskip}{1pt}

\section{Introduction}
Portrait Quality Assessment (PQA) is becoming increasingly important in a variety of fields, from social media engagement to professional photography. The subjective nature of aesthetic appreciation, combined with the technical complexities of image capture and processing, makes PQA a challenging task. While Redi et al.~\cite{redi2015beauty} have explored the attributes that contribute to the perceived beauty of portraits, the utility-focused approach of Face Image Quality Assessment (FIQA)~\cite{schlett2022face} underscores the diversity of criteria required for different quality assessment contexts.

The widespread use of smartphones has democratized portrait photography, yet achieving professional-quality images remains a challenge due to hardware limitations and the intricacies of advanced image processing techniques. Traditional objective quality assessment methods often fall short, as they typically do not account for the non-linear processing involved in modern photography, such as multi-image fusion and AI enhancements~\cite{van2019edge}. This gap has led to the rise of Blind Image Quality Assessment (BIQA) approaches, which evaluate image quality without the need for reference images. However, these methods frequently overlook the scene-specific semantics that significantly influence perceived quality, leading to a ``one-size-fits-all" approach that is rarely effective across varied conditions. The challenges of domain shift and generalization — where the quality assessment model fails to adapt to different conditions — remain as significant obstacles~\cite{zerman2017extensive}.

This paper introduces several novel frameworks aimed at addressing the shortcomings of current PQA methods, particularly in handling domain shifts and ensuring generalizability to unseen portrait conditions. Through an organized challenge, we seek to explore and validate these frameworks, setting new standards for PQA that can adapt to the diverse and dynamic nature of portrait photography. Our challenge relies on the PIQ23 public dataset \cite{Chahine_2023_CVPR} and a private portrait dataset designed specifically to explore the aforementioned challenges.

\paragraph{Related Computer Vision Challenges}
Our challenge is one of the NTIRE 2024 Workshop~\footnote{https://cvlai.net/ntire/2024/} associated challenges on: dense and non-homogeneous dehazing~\cite{ntire2024dehazing}, night photography rendering~\cite{ntire2024night}, blind compressed image enhancement~\cite{ntire2024compressed}, shadow removal~\cite{ntire2024shadow}, efficient super resolution~\cite{ntire2024efficientsr}, image super resolution ($\times$4)~\cite{ntire2024srx4}, light field image super-resolution~\cite{ntire2024lightfield}, stereo image super-resolution~\cite{ntire2024stereosr}, HR depth from images of specular and transparent surfaces~\cite{ntire2024depth}, bracketing image restoration and enhancement~\cite{ntire2024bracketing}, portrait quality assessment~\cite{ntire2024QA_portrait}, quality assessment for AI-generated content~\cite{ntire2024QA_AI}, restore any image model (RAIM) in the wild~\cite{ntire2024raim}, RAW image super-resolution~\cite{ntire2024rawsr}, short-form UGC video quality assessment~\cite{ntire2024QA_UGC}, low light enhancement~\cite{ntire2024lowlight}.


\begin{figure*}[t]
\centering
\includegraphics[width=2\columnwidth]{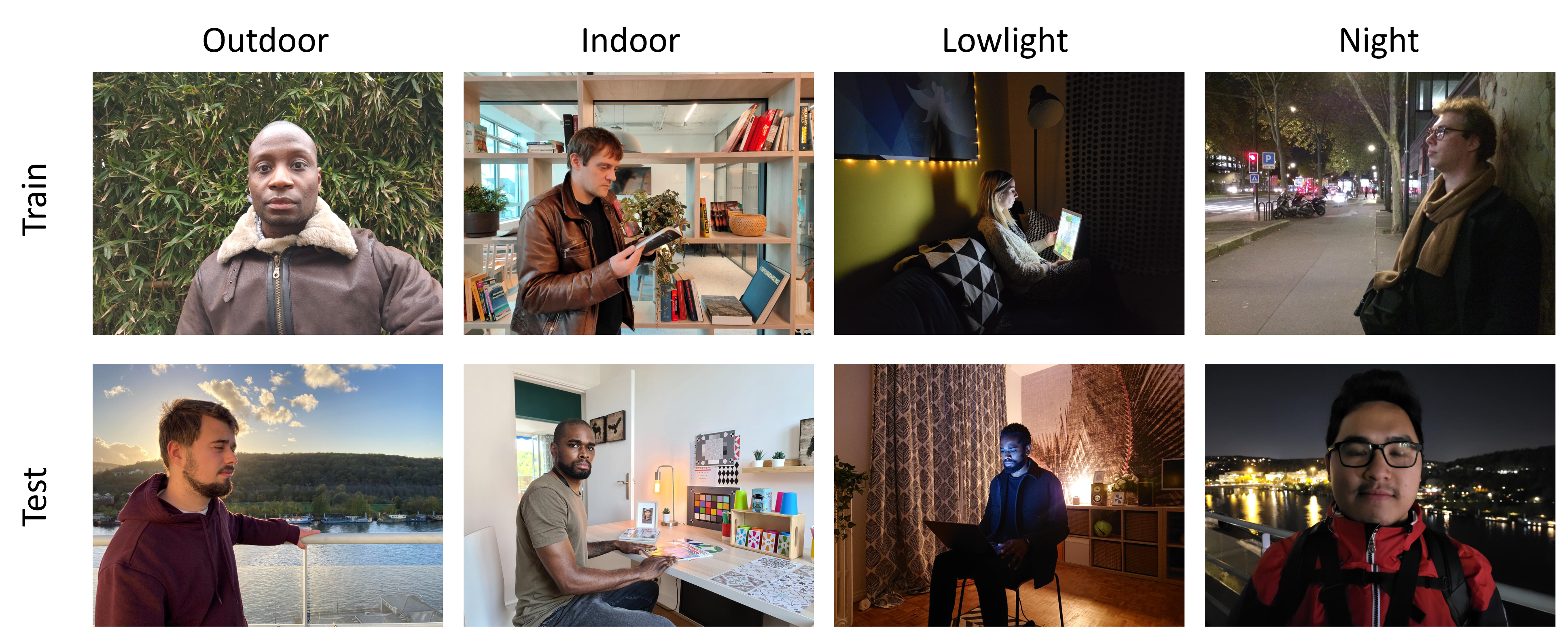}
\setlength{\tabcolsep}{2pt}

\caption{Examples from the \textbf{train/test split of PIQ23~\cite{Chahine_2023_CVPR}}. The test set incorporates various framing settings, backgrounds, subject characteristics, and weather conditions that are significantly distinct from the training set.}
\label{fig:piq23}
\end{figure*}

\begin{figure*}[t]
    \centering
    \begin{subfigure}[b]{0.246\linewidth}
        \includegraphics[width=\linewidth]{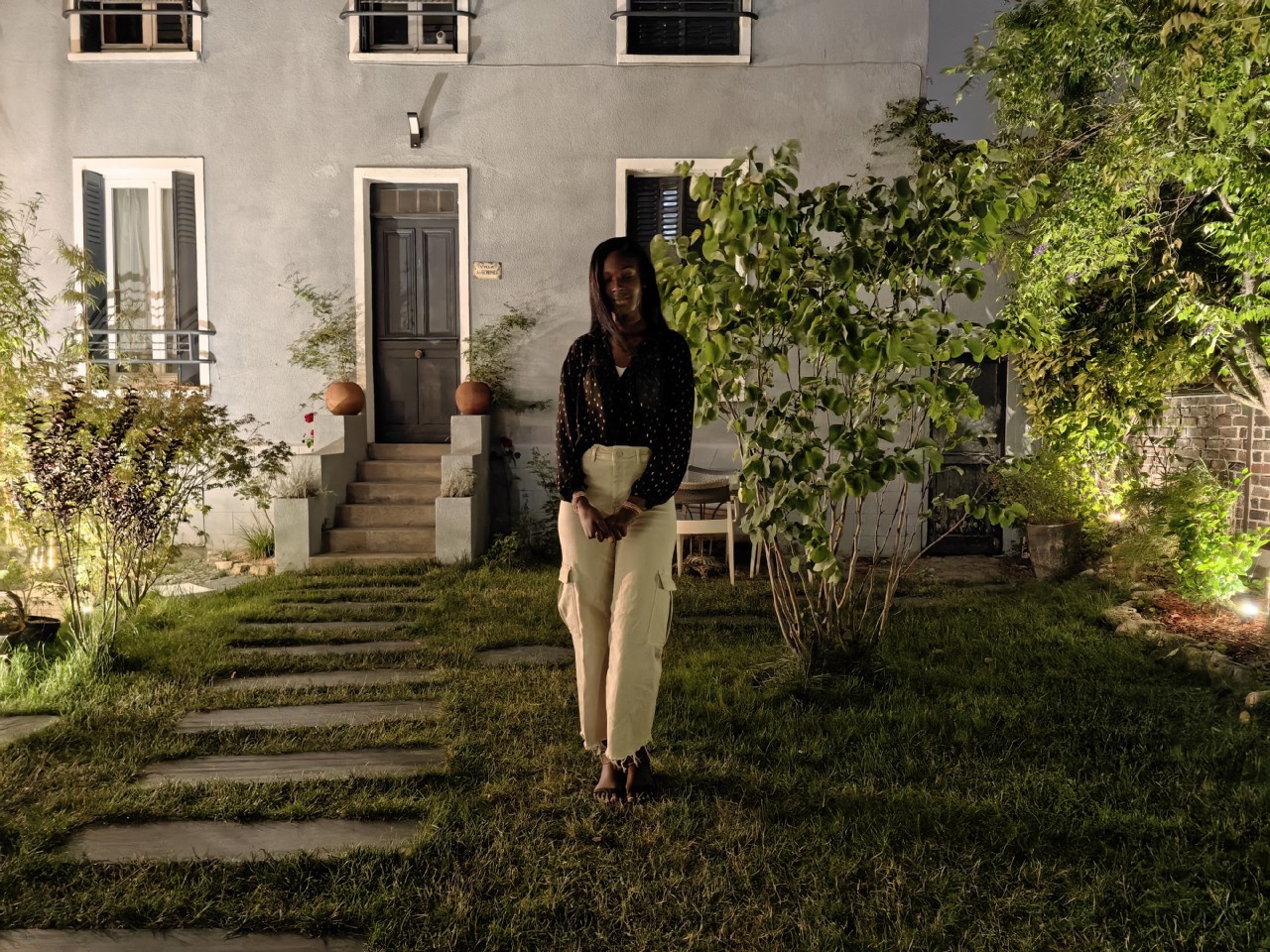}
    \end{subfigure}
    \begin{subfigure}[b]{0.246\linewidth}
        \includegraphics[width=\linewidth]{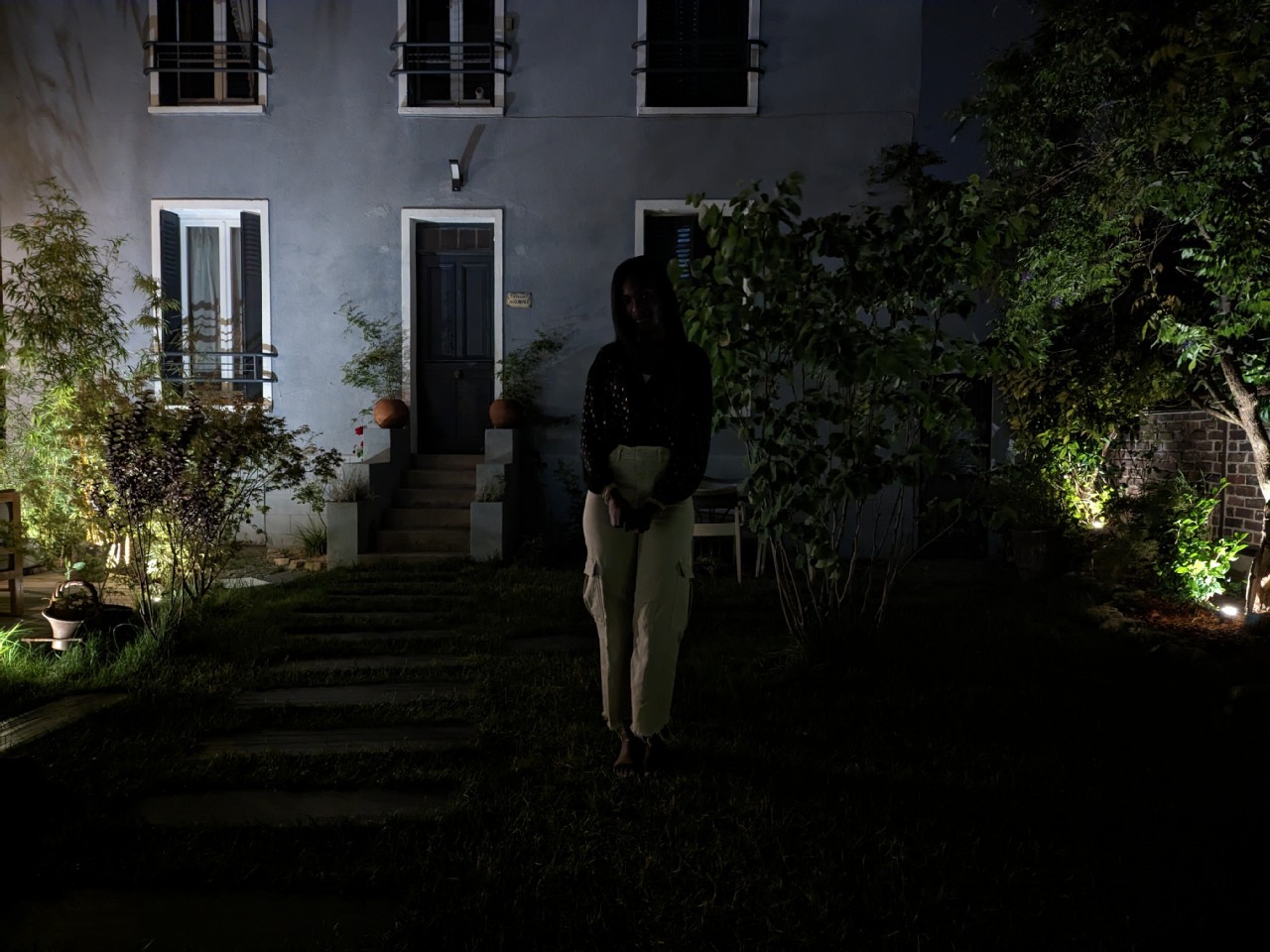}
    \end{subfigure}
    \begin{subfigure}[b]{0.246\linewidth}
        \includegraphics[width=\linewidth]{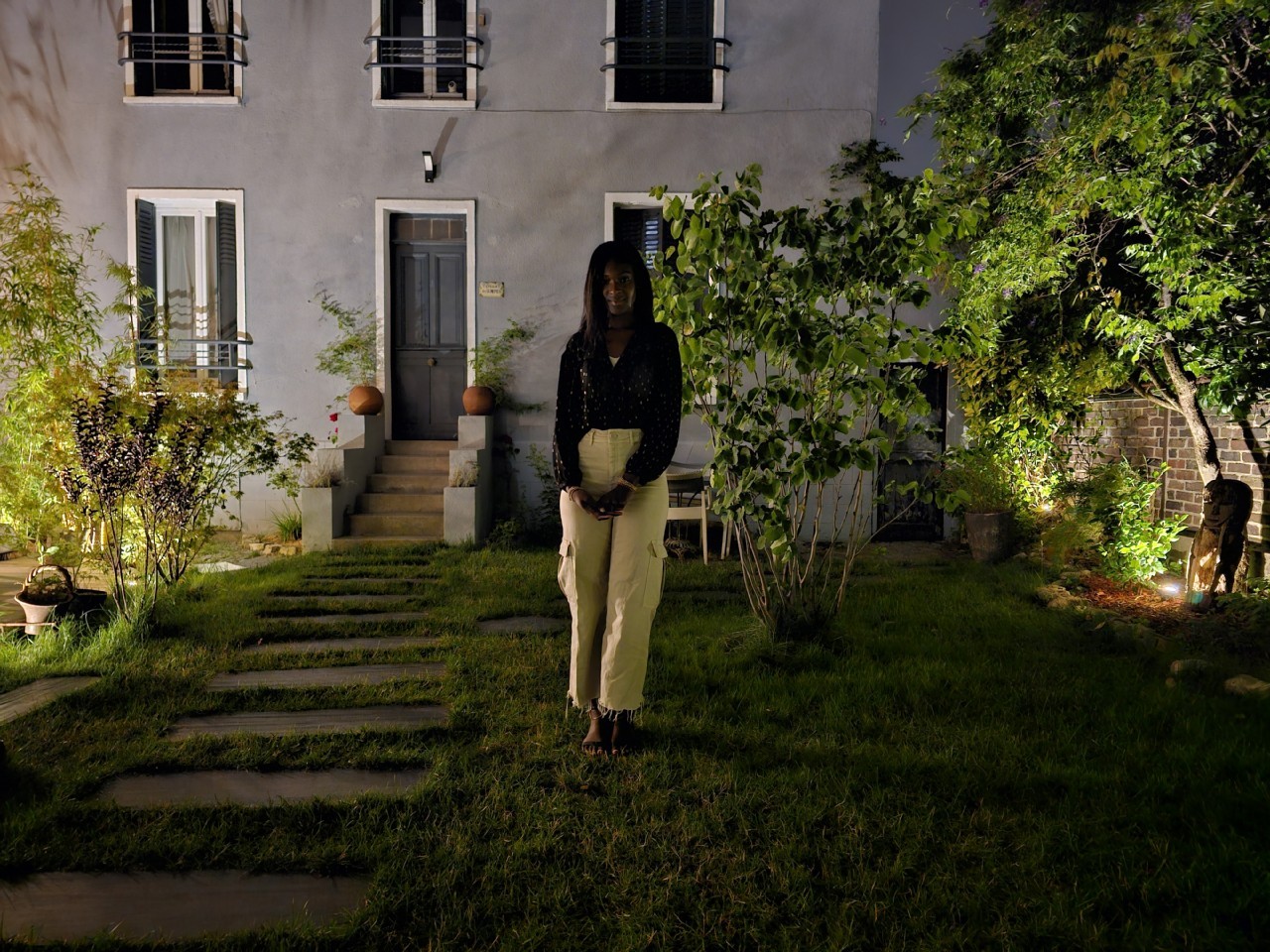}
    \end{subfigure}
    \begin{subfigure}[b]{0.246\linewidth}
        \includegraphics[width=\linewidth]{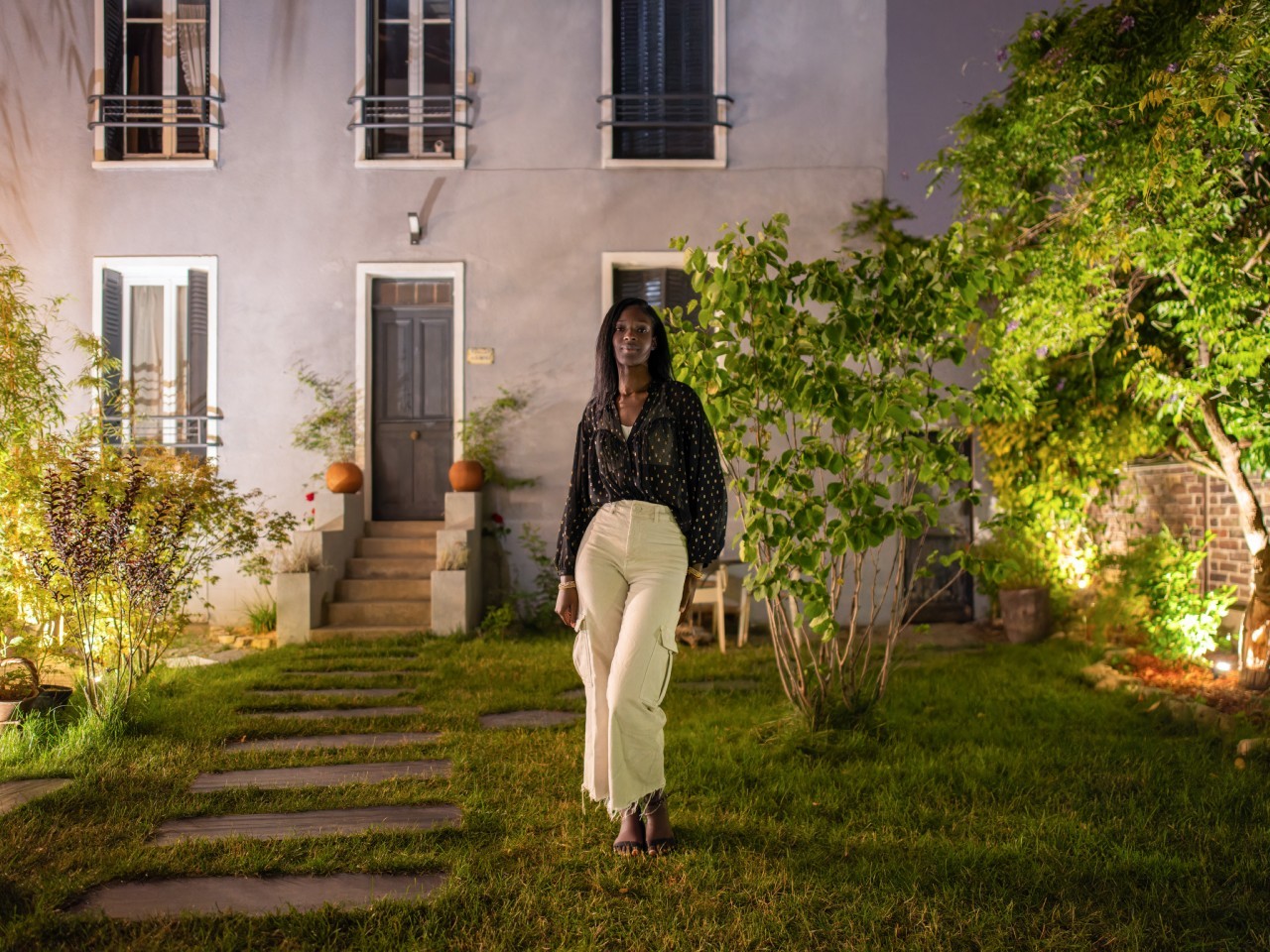}
    \end{subfigure}
    \\
    \begin{subfigure}[b]{0.246\linewidth}
        \includegraphics[width=\linewidth]{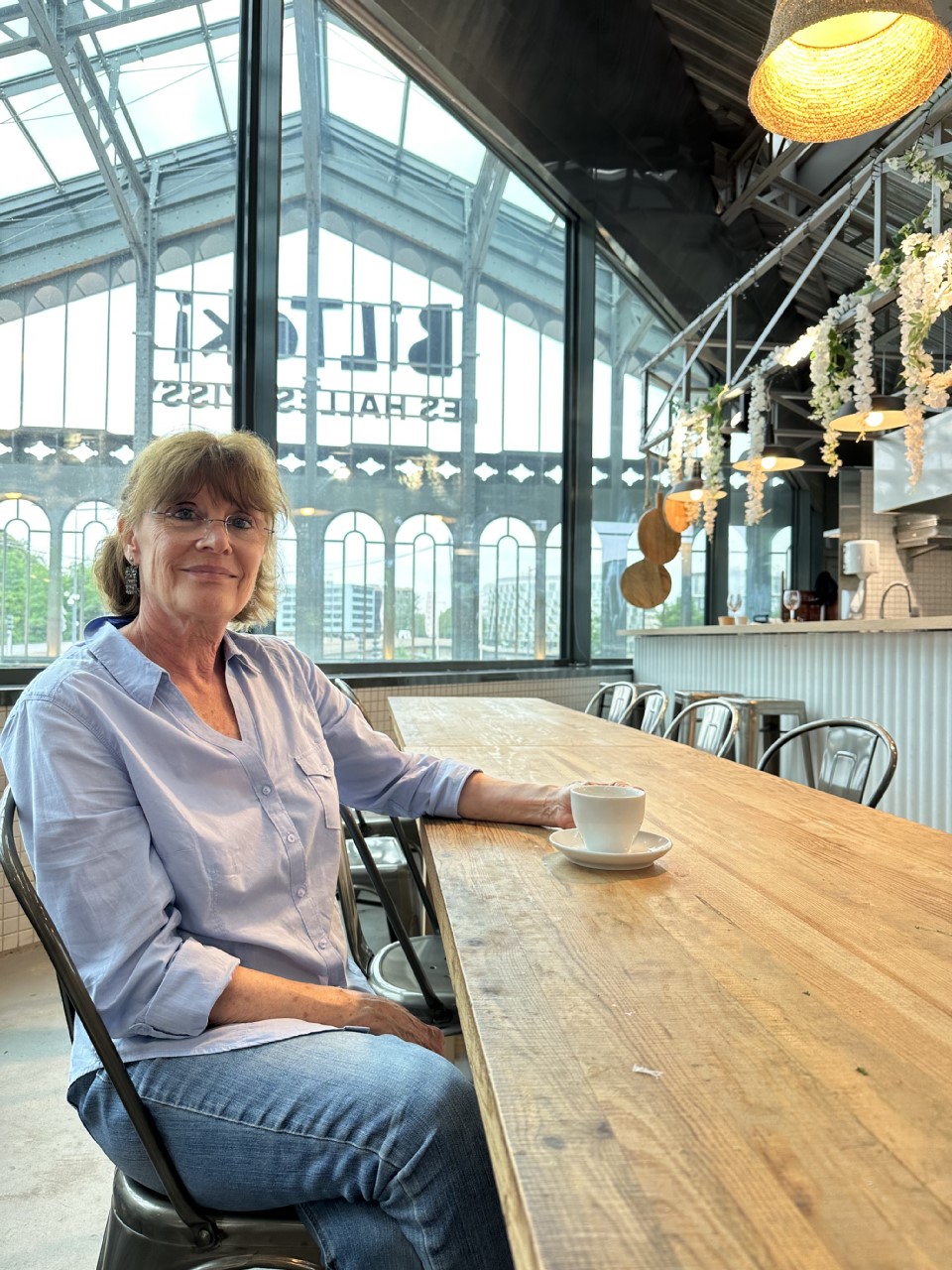}
    \end{subfigure}
    \begin{subfigure}[b]{0.246\linewidth}
        \includegraphics[width=\linewidth]{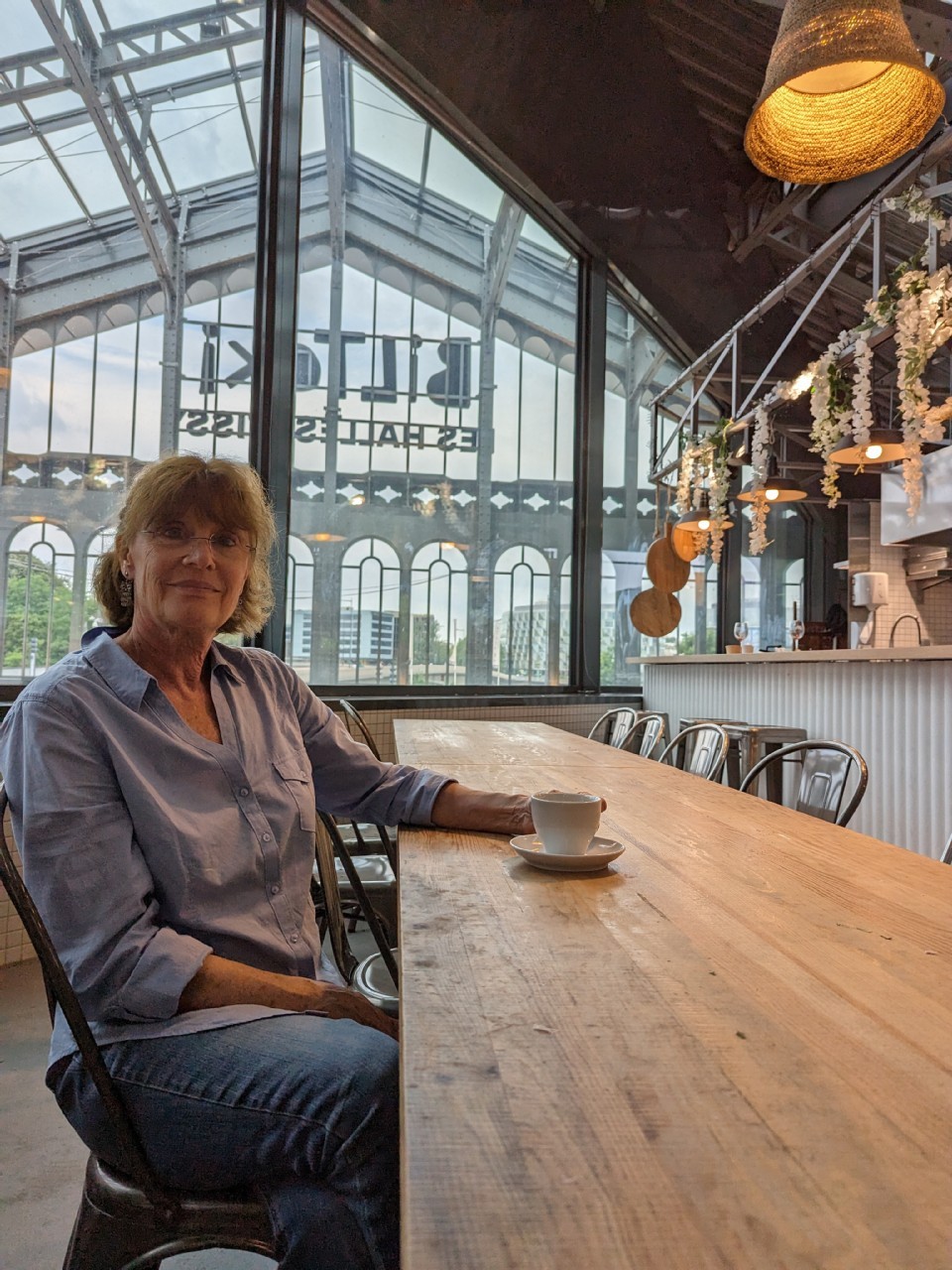}
    \end{subfigure}
    \begin{subfigure}[b]{0.246\linewidth}
        \includegraphics[width=\linewidth]{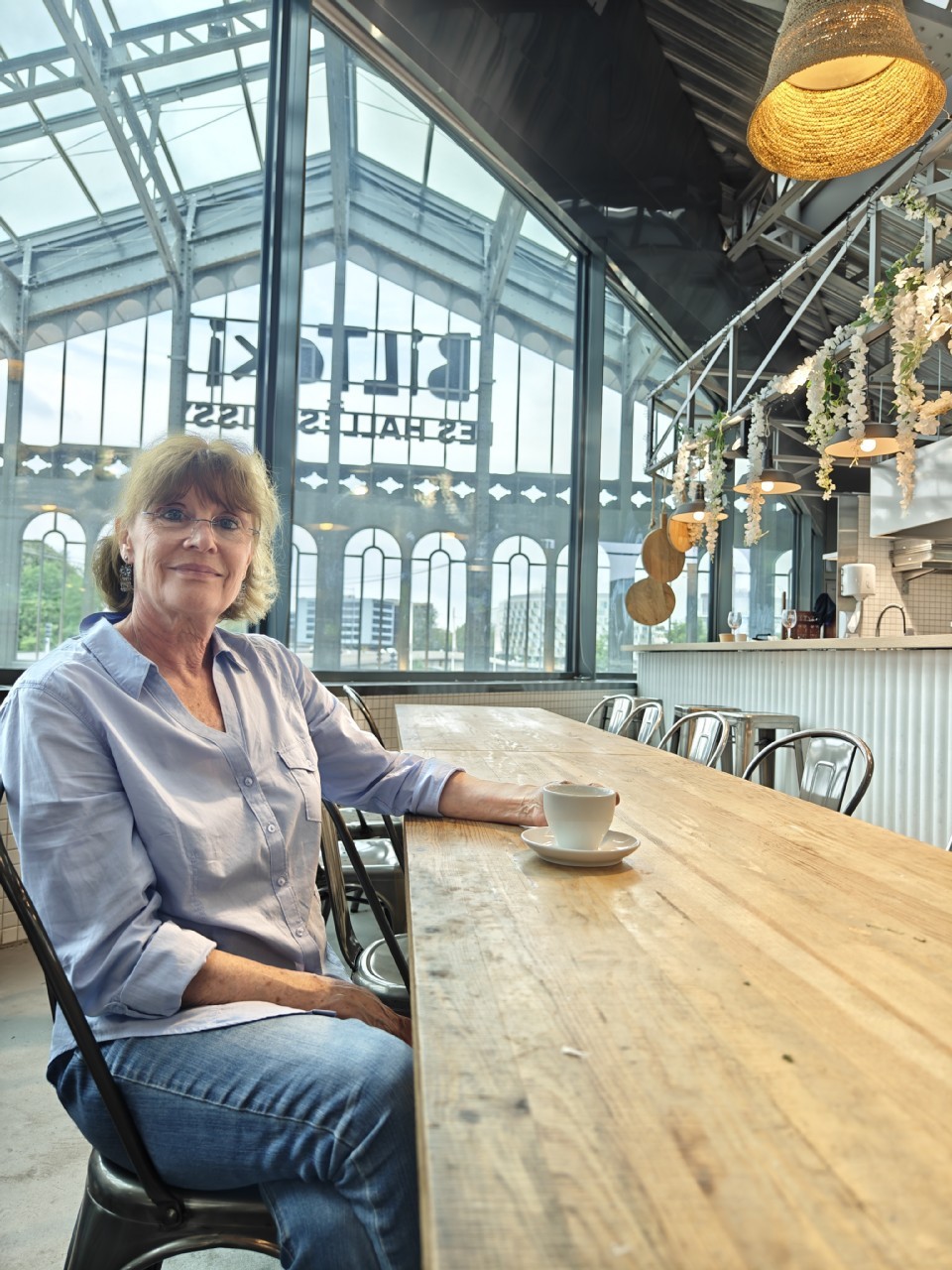}
    \end{subfigure}
    \begin{subfigure}[b]{0.246\linewidth}
        \includegraphics[width=\linewidth]{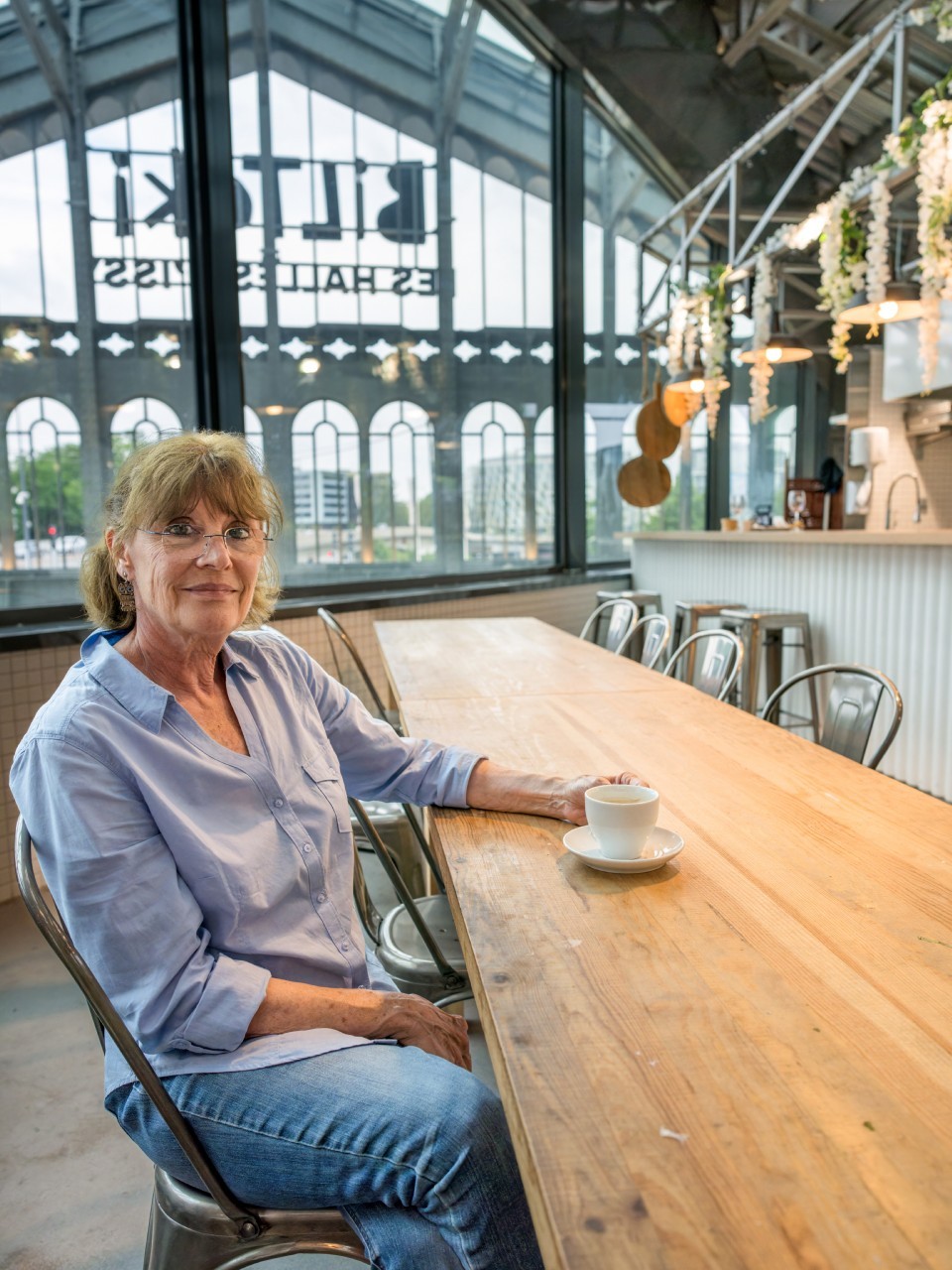}
    \end{subfigure}
    \vspace{-2mm}
    \caption{Sample images from two scenes of the \textbf{challenge generalization test set}. The three first image columns were taken with different smartphone devices, while the last column of images was taken with a DSLR camera and edited by a professional photographer.}
    \label{fig:samples_scene}
\end{figure*}


\begin{figure*}[t]
\centerline{\includegraphics[width=0.8\textwidth]{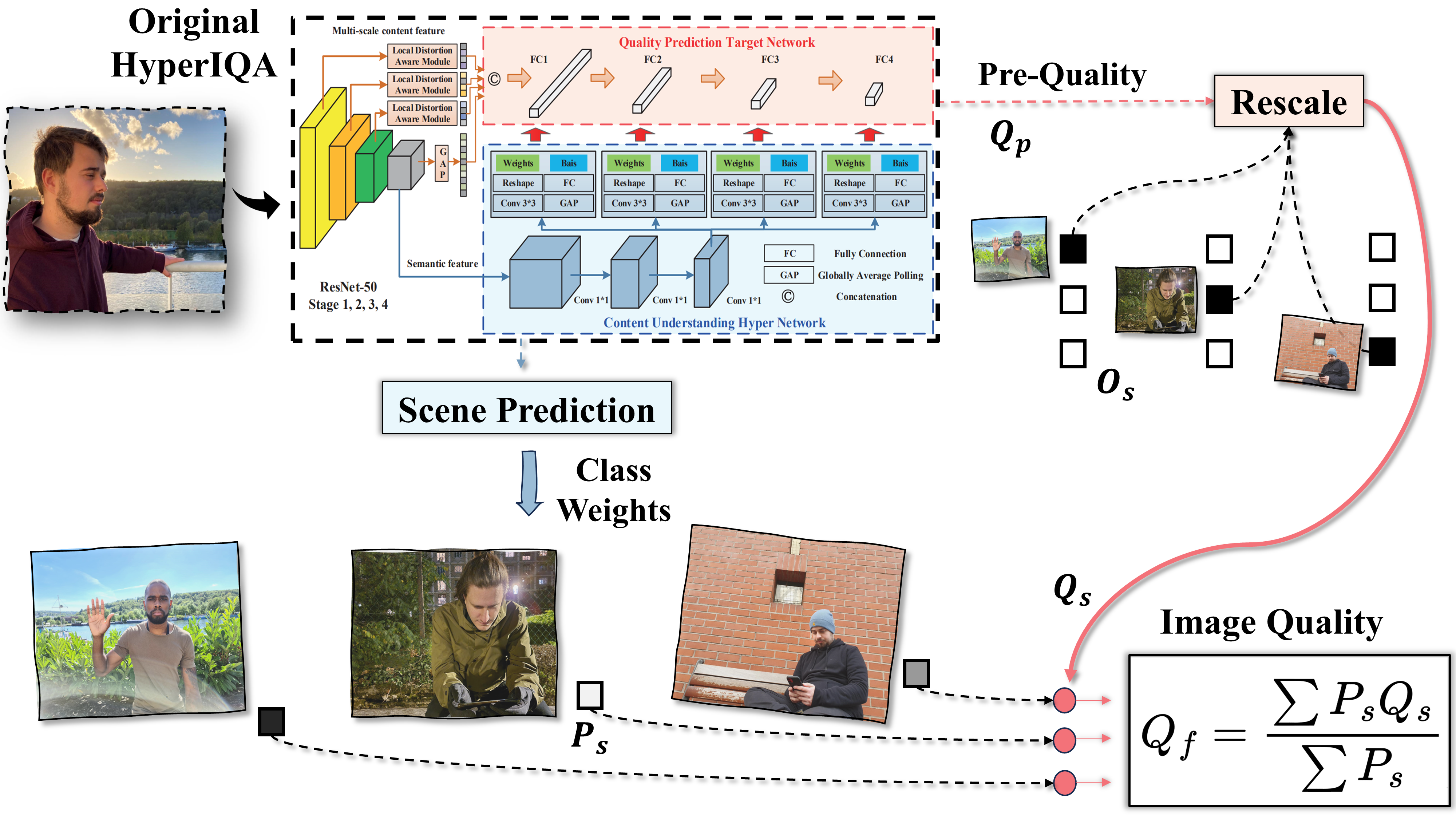}}
\caption{Diagram of FULL-HyperIQA (FHIQA). The figure illustrates how FHIQA processes input images, extracts semantic information, and adapts the quality prediction based on scene-specific evaluations.}
\label{fig:FULL-HyperIQA}
\end{figure*}

\begin{table*}[!ht]
    \centering
    \setlength{\tabcolsep}{5pt}
    \resizebox{0.85\textwidth}{!}{
    \begin{tabular}{l l c c c c c c c}
    \toprule
    & & \multicolumn{3}{c}{PIQ23 Test~\cite{Chahine_2023_CVPR}} & \multicolumn{3}{c}{Challenge Test} \\
    \cline{3-5} \cline{6-8}
    Team & Method & SRCC & PLCC & KRCC & SRCC & PLCC & KRCC \\
    \toprule
    Xidian-IPPL & RQ-Net~(\cref{sec:rqnet}) & 0.820 & 0.839 & 0.621 & \textbf{0.554} & 0.597 & \underline{0.381} \\

    BDVQAGroup  & BDVQA~(\cref{sec:bdvqa})   & \underline{0.849} & \textbf{0.866} & \underline{0.667} & 0.393 & 0.575 & 0.333 \\

    SJTU MMLab  & PQE~(\cref{sec:pqe})   & \textbf{0.864} & \underline{0.857} & \textbf{0.690} & 0.411 & 0.544 & 0.333 \\
    
    I²Group    & MoNet~(\cref{sec:monet})  & 0.760 & 0.791 & 0.580 & 0.357 & 0.433 & 0.286 \\

    SECE-SYSU   & SAR~(\cref{sec:sece}) & 0.828 & 0.855 & 0.651 & 0.304 & 0.453 & 0.238 \\
    
    
    \midrule
    Baseline 1 & HyperIQA~\cite{su2020blindly} & 0.740 & 0.736 & 0.550 & 0.429 & 0.560 & 0.333 \\
    
    Baseline 2 & SEM-HyperIQA~\cite{Chahine_2023_CVPR} & 0.749 & 0.752 & 0.558 & 0.518 & \underline{0.605} & 0.333 \\
    
    Baseline 3 & FULL-HyperIQA~\cite{chahine2024generalized} & 0.778 & 0.784 & 0.586 & \underline{0.536} & \textbf{0.633} & \textbf{0.429} \\
    
    \bottomrule
    \end{tabular}
    }
    \caption{\textbf{Challenge Benchmark.}~SRCC: Spearman Rank Correlation Coefficient, PLCC: Pearson Linear Correlation Coefficient, KRCC: Kendal Rank Correlation Coefficient.
    The correlation metrics are calculated per scene, and the final result corresponds to the median of scene-wise metrics. We highlight the \textbf{best} and \underline{second best}.
    }
    \label{tab:benchmark}
\end{table*}

\begin{table*}[!ht]
    \centering
    \setlength{\tabcolsep}{4pt}
    \resizebox{0.99\textwidth}{!}{
    \begin{tabular}{l l c c c c c c}
    \toprule
    Team & Method & PIQ23 Test~\cite{Chahine_2023_CVPR} & Challenge Test & \# Params. (M) & Extra Data & Train Res. \\
    \toprule
    Xidian-IPPL & RQ-Net~(\cref{sec:rqnet}) & 0.751 & \textbf{0.517} & \textbf{57.6} & Yes & 224 \\

    BDVQAGroup  & BDVQA~(\cref{sec:bdvqa})  & \underline{0.779} & 0.433 & 794 & No & 384  \\

    SJTU MMLab  & PQE~(\cref{sec:pqe})  & \textbf{0.811} & 0.429 & 174 & Yes & 384  \\
    
    I²Group     & MoNet~(\cref{sec:monet})  & 0.710 & 0.368  & 408 & No & 384  \\

    SECE-SYSU   & SAR~(\cref{sec:sece})    & 0.777 & 0.315  & 149  & No & 224  \\
    
    
    \midrule
    Baseline 1 & HyperIQA~\cite{su2020blindly} & 0.676 & 0.456 & \underline{128} & No & 1300px \\
    Baseline 2 & SEM-HyperIQA~\cite{Chahine_2023_CVPR} &0.690  & 0.501 & 145 & No & 1300px \\
    Baseline 3 & FULL-HyperIQA~\cite{chahine2024generalized} & 0.711 & \underline{0.515} & 145 & No & 1300px \\
    
    \bottomrule
    \end{tabular}
    }
    \caption{The final metric for each testing set consists of the median of the scene-wise average of the SRCC, PLCC, and KRCC correlations. We also provide the training resolution in pixels (px), number of parameters (in Millions), and if the team used additional data for training.}
    \label{tab:details}
\end{table*}

\section{{Portrait IQA Challenge}}

In this challenge, we introduce the PIQ benchmark \cite{ntire2024QA_portrait}, based on the PIQ23 portrait dataset \cite{Chahine_2023_CVPR} published in 2023, composed of diverse skin tone photographs in challenging scenarios for smartphone cameras. The dataset is divided into 50 “scenes” defined by their illumination condition, target distance, framing, posture, background, etc. Every scene has around 100 images collected from multiple smartphones and covering various subjects. Each scene is separately annotated according to three image quality attributes  (detail/noise, exposure/contrast, and overall) using pairwise comparisons, which yields precise and consistent quality insights when applied to image groups with similar content. Around 600k comparisons in total (for the 3 features) were collected from 30 experts in controlled visualization conditions (calibrated screens, fixed eye-to-screen distance, controlled background illumination, etc. \cite{Chahine_2023_CVPR}). These annotations were converted to JODs (Just Objectionable Difference), quality units where 1 unit apart means that 75\% of the observers can see the quality difference between two images, using psychometric scaling algorithms. 
By design, each scene has an independent quality scale where the scores of the scenes are not inter-comparable. This introduces a challenge when training machine learning models. 

\paragraph{Test Dataset and Evaluation}
In this challenge, we proposed to focus on the \emph{overall} attribute and a \emph{``generalization split”} (we will refer to this as the challenge testing set, hidden/private test), that is, to evaluate the capacity of the models to generalize outside the training scenes when evaluating the overall quality of the portrait. We cannot expect the ML model to correctly estimate the JOD quality value of the image since it is dependent on the scene, but it should be able to correctly rank a set of images according to their quality. 

We split the \textbf{evaluation procedure} into two phases. For the preliminary testing phase, we propose to use the public PIQ23 (\cref{fig:piq23}) test set with no scene overlap with the training set (the images of scene 1 in the training set cannot be in the testing set). The final testing phase is based on a private test set composed of 96 single-person scenes of 7 images each, taken with 6 high-quality smartphone images and 1 DSLR capture edited by a photographer used as the quality reference (\cref{fig:samples,fig:samples_scene}). 

The participants do not have access to the challenge generalization testset. The results are obtained by executing their submitted models to ensure reproducibility, and basic runtime requirements on commercial GPUs.

\subsection{Baseline Models}

We have chosen to compare the proposed models with multiple baseline methods from the HyperIQA family (HyperIQA \cite{ su2020blindly}, SEM-HyperIQA \cite{Chahine_2023_CVPR} and FHIQA \cite{chahine2024generalized}) which are specifically designed to tackle the domain shift and scene semantics understanding, and that have proven performance on the PIQ23 dataset. 

HyperIQA uses the HyperNetwork architecture to incorporate semantic information into image quality predictions. Building upon HyperIQA, SEM-HyperIQA introduces a multitasking approach that allows for scene-specific rescaling of quality scores. It employs a multi-layer perceptron (MLP) to predict the scene category of an image, which is then used to adjust the quality scores of individual patches through a scene-specific multiplier and offset. However, it assumes that each image belongs to a known scene category, which limits its ability to generalize to new scenes. FHIQA (\cref{fig:FULL-HyperIQA}) extends the concept of quality score rescaling by utilizing the entire scene prediction vector, rather than relying on a single scene category. This vector represents the similarity of the input image to all known scene categories, allowing FHIQA to rescale the pre-quality score based on a weighted combination of these similarities. The key innovation of FHIQA lies in its potential to generalize to new scene categories not included in the training set, by leveraging the information encoded in the classification weights.

\newpage

\section{Challenge Results and Methods}
\label{sec:teams}

In the following sections we describe the best challenge solutions. Note that the method descriptions were provided by each team as their contribution to this survey. 

In \cref{tab:benchmark} we provide the benchmark using standard correlation metrics. We can observe that all the methods struggle to generalize in the challenge testset. The reason is that the new test images were captured using high-quality smartphones, extending the PIQ23~\cite{Chahine_2023_CVPR} dataset. The models struggle with this quality domain gap, which indicates that the model performance highly depends on the device used for capturing the data. 

In \cref{tab:details} we provide the final ranking and additional information of the methods.

\newpage
\subsection{RQ-Net: Towards Robust Cross-scene Relative Quality Assessment}
\label{sec:rqnet}


\begin{center}

\vspace{2mm}
\noindent\emph{\textbf{Team Xidian IPPL}}
\vspace{2mm}

\noindent\emph{Zhichao Duan,
Xinrui Xu,
Yipo Huang,
Quan Yuan,
Xiangfei Sheng,
Zhichao Yang,
Leida Li \\}

\vspace{2mm}

\noindent\emph{
Xidian University}

\vspace{2mm}

\noindent{\emph{Contact: \url{zach@stu.xidian.edu.cn}}}

\end{center}

The team presents a method for Robust Cross-scene \textbf{R}elative \textbf{Q}uality Assessment.

\begin{figure*}[t]
    \centering
    \includegraphics[width=\linewidth]{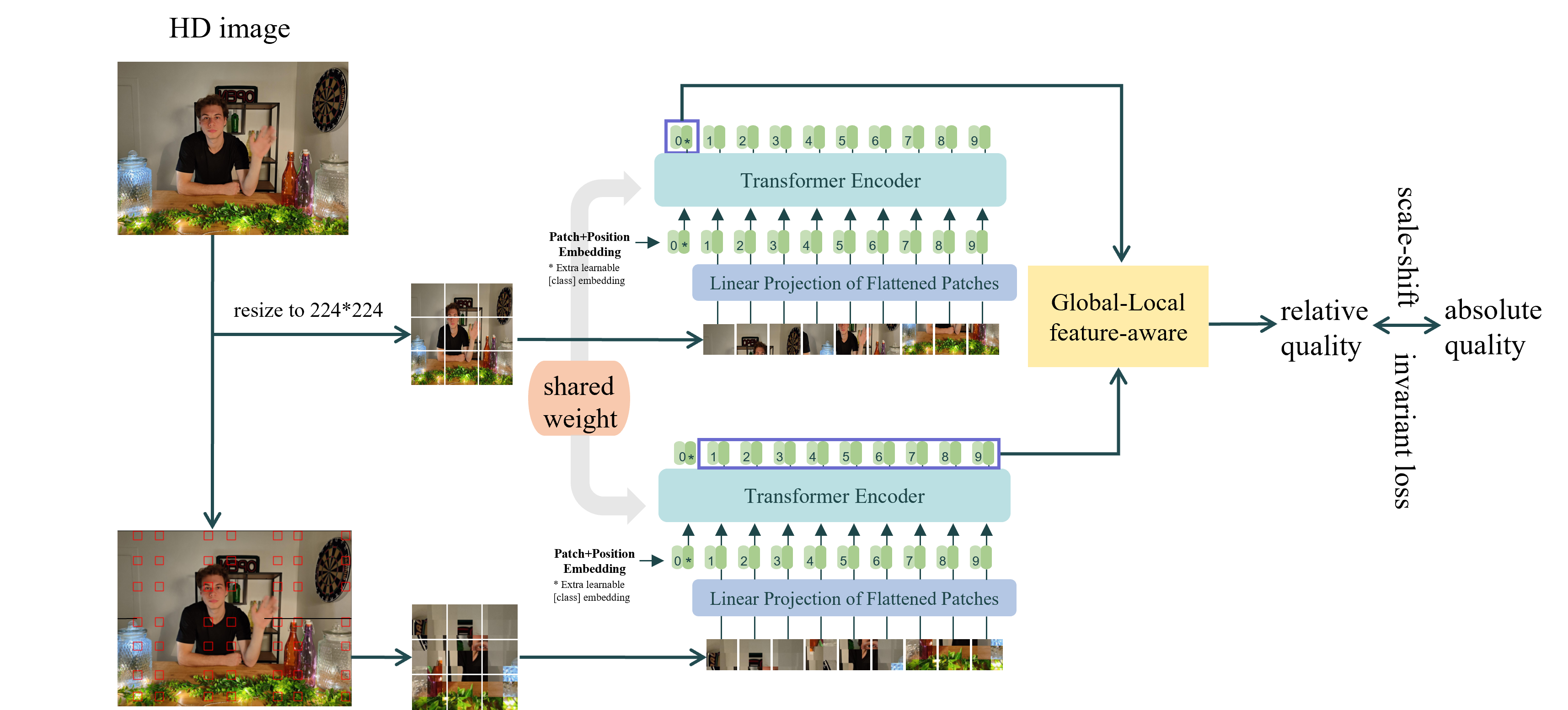}
    \caption{Diagram of the \textbf{RQ-Net} proposed by Team Xidian IPPL.}
    \label{fig:team4}
\end{figure*}

RQ-Net is a method to predict the relative quality of images. It consists of two branches: global quality perception and local quality perception. As shown in \cref{fig:team4}. A downsampled version of image used as input to the global branch, and multiple patches cropped from HD image are used as input to the local branch. This simple design is inspired by some previous work~\cite{9536693,wu2022fast}. Both branches use ViT-B/16~\cite{DBLP:journals/corr/abs-2010-11929} with shared weights as the backbone and are initialized with CLIP~\cite{clip} pre-trained weights. After ViT encoding, the class (global) features and grid (local) features are fused by a Global-Local Feature-aware Block and the relative quality scores of the images are predicted. We propose the following two main contributions to achieve Robust Cross-scene ``Relative Quality'' Assessment: 

\vspace{2mm}

(1) \textbf{Scale-shift invariant loss.} Attempting to train with data from different scenes/domain for cross-scene generalization is difficult and inappropriate due to the different quality label scales of images between scenes and the presence of domain shift. 
We propose to predict quality in a ``relative quality space'' with scale-shift invariant loss to handle this ambiguity. In a mini-batch prediction, let $S$ be the number of scene categories in the batch and $K$ be the sample size of each category. Then we define the scale-shift invariant loss as:
\begin{equation}
    \mathcal{L} = \frac{1}{SK}\sum_{i=1}^{S}\sum_{j=1}^{K}\left\|\hat{q}_{ij}-\hat{q}_{ij}^*\right\|
\end{equation}
where $\hat{q}_{ij}$ and $\hat{q}_{ij}^*$ are the scores of the prediction and ground truth after mapping them into relative quality space. For $K$ samples in each scene, we use a simple and robust way to map predictions and ground truth to a zero-shift and unit-scaled quality space:
\begin{equation}
\begin{aligned}
    t(q)=median(q),& \quad s(q)=mean(\left\|q-t(q)\right\|) \\
    \hat{q} = \frac{q-t(q)}{s(q)},& \quad \hat{q}^*=\frac{q^*-t(q*)}{s(q^*)}
\end{aligned}
\end{equation}
We uses the parameters $S, K$ and a custom \verb|torch.utils.data.Sampler| to balance the scene richness and sample richness of the training process.

\vspace{2mm}
(2) \textbf{Pre-training with mixed multi-source data.} 
Since we train the model in relative quality space, multiple datasets can be easily blended for joint tuning. We propose two pre-training strategies and use bagging ensemble method, RQ\textsubscript{\textit{general}} and RQ\textsubscript{\textit{portrait}} are trained for ensembles.
Specifically, we mix four datasets SPAQ~\cite{fang2020perceptual}, KonIQ-10k~\cite{hosu2020koniq}, LIVE In the Wild~\cite{ghadiyaram2015massive} and RBID~\cite{ciancio2010no} and consider them to be from four different scenes (domains). RQ\textsubscript{\textit{general}} is pre-trained in the relative quality space and fine-tuned on the PIQ23.
In addition, we construct the PIQ23-Face dataset by masking the regions outside the face in the PIQ23 image. RQ\textsubscript{\textit{portrait}} is pretrained on PIQ23-Face with the same strategy, but using three separate Global-Local Feature-aware Blocks to perceive detail, exposure and overall quality. Finally fine-tuned on PIQ23.
The two pre-training approaches greatly promote the model's cross-scene evaluation capability and the robustness of portrait evaluation. The model is illustrated in \cref{fig:team4}.

\vspace{2mm}

We use ViT-B/16~\cite{DBLP:journals/corr/abs-2010-11929} as the backbone and are initialized with CLIP~\cite{clip} \textbf{pre-trained} weights.

We pre-train using two types of datasets and then fine-tune on the PIQ23-Overall provided by the challenge. The data used for pre-training is:\newline
(1) External public datasets: SPAQ~\cite{fang2020perceptual}, KonIQ-10k~\cite{hosu2020koniq}, LIVE In the Wild~\cite{ghadiyaram2015massive} and RBID~\cite{ciancio2010no}. \newline
(2) PIQ23-Face: Obtained by pre-processing the PIQ23 dataset. Regions other than the face in the original image are masked to 0, and the detail, exposure and overall quality scores in the original dataset are used for supervised multi-task learning.

\vspace{2mm}

\noindent\textbf{Training:} The original HD image is first resized to 244$\times$244, then randomly crop and flip for augmentation. The cropped image of size 224$\times$224 is used as input for the global branch. For the local branch, the HD image is also applied with random flip and then divided into 7$\times$7$=$49 squares. The 32$\times$32 sized mini-patches cropped from each square are re-spliced into 224$\times$224 sized inputs. 

\vspace{2mm}

\noindent\textbf{Inference:} The inputs for testing and training are the same. But for testing augmentation, four-corner, top, bottom, left, right, and center crops are used instead of randomly flipping and cropping the images.

\paragraph{Implementation details}

The team implemented RQ-Net by PyTorch and train it on two NVIDIA 4090 GPUs. The original HD image is first resized to 244$\times$244, then randomly crop and flip for augmentation. The cropped image of size 224$\times$224 is used as input for the global branch. For the local branch, the HD image is also applied with random flip and then divided into 7$\times$7$=$49 squares. The 32$\times$32 sized mini-patches cropped from each square are re-spliced into 224$\times$224 sized inputs. The inputs for testing and training are the same. But for testing augmentation, four-corner, top, bottom, left, right, and center crops are used instead of randomly flipping and cropping the images.

We use the Adam optimizer with weight decay of $1\times10^{-5}$ to train RQ-Net, with mini-batch size of 128 ($S$=4, $K$=32). We use the cosine decay learning rate strategy, with a maximum learning rate of $1\times10^{-5}$.

Four models with the same structure were eventually trained for ensemble. RQ\textsubscript{\textit{general-M1}}, RQ\textsubscript{\textit{portrait-M1}}, RQ\textsubscript{\textit{general-M2}} and RQ\textsubscript{\textit{portrait-M2}}, where general/portrait denotes the two pre-training strategies introduced previously, and M1/M2 denotes different training set division strategies.

\subsection{Ranking based vision transformer network for image quality assessment.}
\label{sec:bdvqa}


\begin{center}

\vspace{2mm}
\noindent\emph{\textbf{Team BDVQA Group}}
\vspace{2mm}

\noindent\emph{Haotian Fan,
Fangyuan Kong,
Yifang Xu
}

\vspace{2mm}

\noindent\emph{ByteDance Inc}

\end{center}

The team proposed a method based on MSTRIQ~\citet{wang2022mstriq}, a Swin-Transformer based method. We raise several training and inference tricks to increase the performance of this method. We combined rank loss and mse loss to increase the model same-scene ranking ability.

The merged ranking loss is given by:                       
\begin{equation}
\resizebox{.9\hsize}{!}{$
loss_{merged\_loss} = \frac{2}{N}\sum_{i=0:2:N}
\begin{cases}
e^{\hat{y}^i-\hat{y}^{i+1}}+ (y^i -\hat{y})^{2} , & \text{if } { y^i<y^{i+1}} \\
(y^i -\hat{y})^{2},  & \text{others } 
\end{cases}
$}
\end{equation}

We used several data augmentation method to increase the training dataset and enhance robustness of our model: Random Crop image into patches, Random Rotation.
  
We also use \emph{Test time augmentation (TTA)} can perform random modifications to the testing images. The following TTA methods are implemented to increase our model performance: (i) FiveCrop and TenCrop, (ii) Random crop then inference each image 18 times averaged.

\begin{figure}[t]
    \centering
    \includegraphics[width=\linewidth]{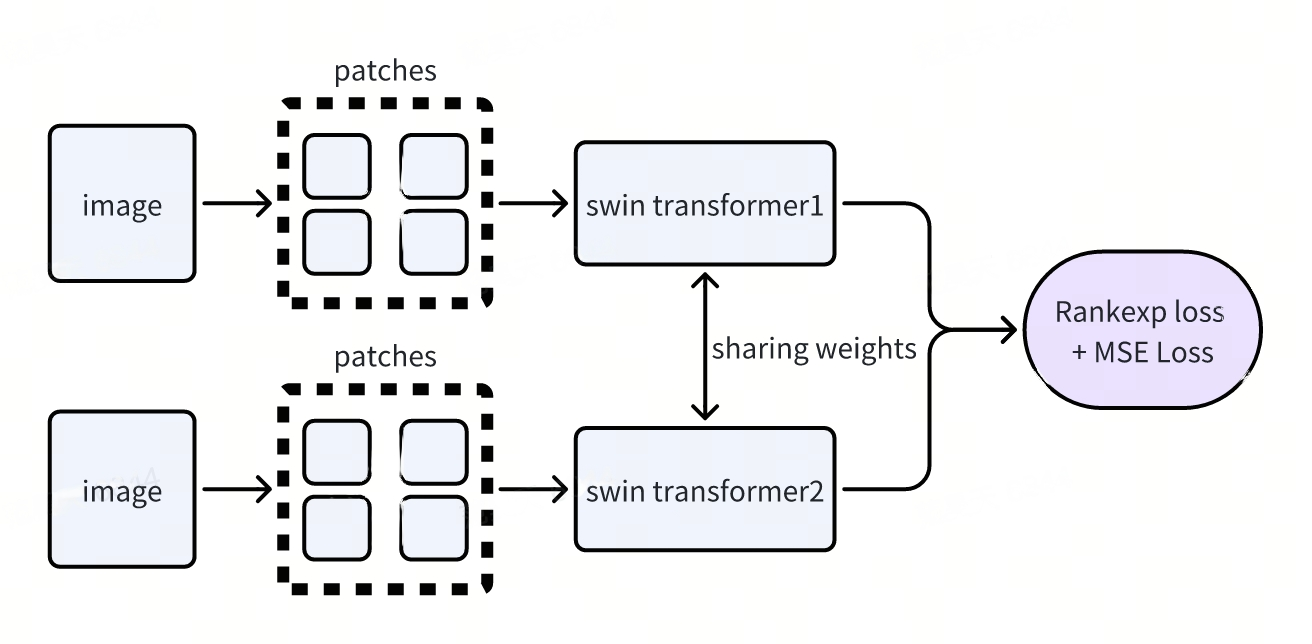}
    \caption{Siamese Swin Transformer~\cite{liu2021swin} approach proposed by Team BDVQA.}
    \label{fig:bdvqa}
\end{figure}

We use Swin transformer~\cite{liu2021swin} pre-trained on ImageNet. No additional datasets were used. The data pre-processing consists on random resized crops. The overall training method is illustrated in \cref{fig:bdvqa}.

\paragraph{Implementation details}

The model is implemented in Pytorch. The estimated training time is 2h using 8 A100 GPUs (40G). The models are trained using AdamW optimizer and learning rate $2e^{-5}$.

The input images are augmented using random resized crop to 384 x 384.
During inference, we use test time augmentation (TTA) of random crops 18 times, and average the results to produce the final output.

\begin{figure*}[t]
    \centering
    \includegraphics[trim={1cm 2cm 1cm 2cm},clip, width=\linewidth]{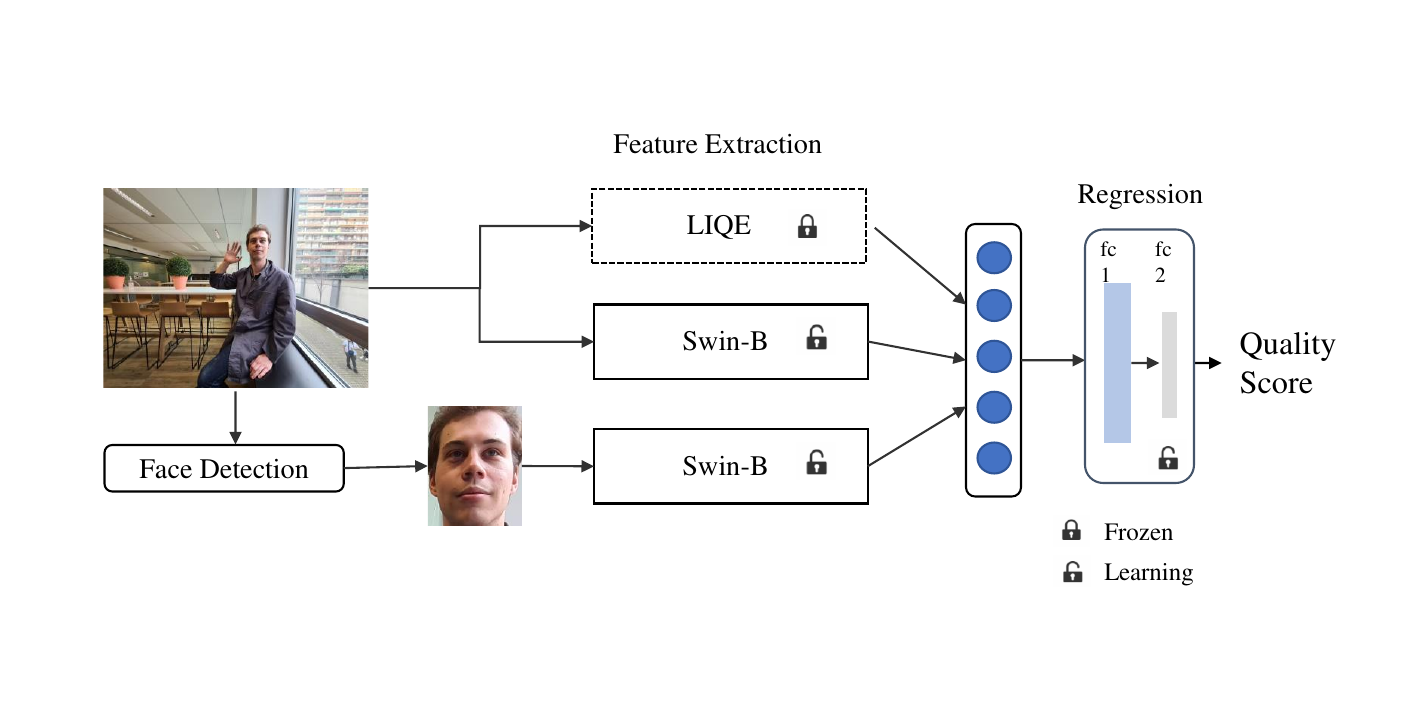}
    \caption{PQE method proposed by Team SJTU MMLab.}
    \label{fig:team3}
\end{figure*}

\subsection{PQE: A Portrait Quality Evaluator by Analyzing the Characteristics of Facial and Full Images}
\label{sec:pqe}


\begin{center}

\vspace{2mm}
\noindent\emph{\textbf{Team SJTU MMLab}}
\vspace{2mm}

\noindent\emph{Wei Sun~$^1$,
Weixia Zhang~$^1$,
Yanwei Jiang~$^1$,
Haoning Wu~$^2$,
Zicheng Zhang~$^1$,
Jun Jia~$^1$,
Yingjie Zhou~$^1$,
Zhongpeng Ji~$^3$,
Xiongkuo Min~$^1$,
Weisi Lin~$^2$,
Guangtao Zhai~$^1$
}

\vspace{2mm}

\noindent\emph{
$^1$ Shanghai Jiao Tong University\\
$^2$ Nanyang Technological University\\
$^3$ Huawei
}

\vspace{2mm}

\noindent{\emph{Contact: \url{sunguwei@sjtu.edu.cn}}}

\end{center}

The team introduces a two-branch portrait quality assessment model motivated by the influence of both facial and background components on portrait quality. Thus, employing a single neural network on the portrait image is insufficient to model the quality relationship between the facial and the background components. 

To address this problem, we propose a two-branch neural network (each branch consisting of a Swin Transformer-B~\cite{liu2021swin}) for portrait quality assessment, where two branches are used to model the quality characteristics of the full and the facial components respectively. 

Moreover, the shooting scene (including luminance, environment, etc.) also impact the perception of portrait quality. Therefore, we perform LIQE~\cite{zhang2023blind}, a CLIP based scene classification and quality evaluation model, to extract scene and quality features for the full image. Subsequently, we concatenate these features and utilize a two-layer MLP to derive the quality scores. We employ the learning-to-rank training method~\cite{zhang2023blind} and use the fidelity loss \cite{tsai2007frank} as the loss function to optimize the model.

We use LIQE, a \textbf{pre-trained} model trained on LIVE~\cite{sheikh2006statistical}, CSIQ~\cite{larson2010most}, KADID-10k~\cite{lin2019kadid}, BID~\cite{ciancio2010no}, CLIVE~\cite{ghadiyaram2015massive}, and KonIQ-10k~\cite{hosu2020koniq} to extract scene and quality features. 
The branch for the entire image is pre-trained on the LSVQ~\cite{ying2021patch} dataset and the branch for the facial image is pre-trained on the GFIQA~\cite{su2023going} dataset. For images in the PIQ23 dataset, we use yolo-face package to extract the face images from the full portrait image.

\vspace{2mm}
\noindent \textbf{Training:} We use the fidelity loss to train our model. Specifically, for an image pair ($\bm{x}$, $\bm{y}$) from the same scene in the PIQ23 dataset, we compute a binary label according to their ground-truth JODs:

\begin{equation}
    p(\bm{x}, \bm{y}) = \left\{ 
    \begin{aligned} 
    & 0 \quad \rm{if} \quad q(\bm{x}) \geq q(\bm{y}))\\
    & 1 \quad \rm{otherwise}
    \end{aligned}
    \right.
\end{equation}

\noindent We estimate the probability of $\bm{x}$ perceived better than $\bm{y}$ as

\begin{equation}
\hat{p}(\bm{x}, \bm{y}) = \Phi(\frac{\hat{q}(\bm{x})-\hat{q}(\bm{y})}{\sqrt{2}}),
\end{equation}
where $\Phi(\cdot)$ is the standard Normal cumulative distribution function, and the variance is fixed to one. We adopt the fidelity loss to optimize the model:
\begin{equation}
\begin{aligned}
\ell(\bm{x}, \bm{y};\bm{\theta}) = &1-\sqrt{p(\bm{x}, \bm{y})\hat{p}(\bm{x}, \bm{y})}\\
&-\sqrt{(1-p(\bm{x}, \bm{y}))(1-\hat{p}(\bm{x}, \bm{y}))}.
\end{aligned}
\end{equation}

\vspace{1mm}

During training, both the resolutions of full and facial images are resized to $384\times384$. We train the model on 2 NVIDIA RTX 3090 GPUs with a batch size of 6. The training epoches are set as 10. Learning rate is $1\times10^{-5}$.

We use the Swin Transformer-B as the backbone, which is a hybrid network structure. We use yolo-face as the face detector, which is a CNN network structure. LIQE is the transformer based network structure. The model is illustrated \cref{fig:team3}.

\paragraph{Implementation details}

\begin{itemize}
\item Optimizer: Adam
\item Learning rate: $1\times10^{-5}$
\item GPUS: 2 NVIDIA RTX 3090
\item Datasets: We use the LSVQ dataset to pre-train the branch for the full image to obtain a robust quality-aware feature representation and use the GFIQA dataset to pre-train the branch for the facial image to obtain facial-related quality feature representation. We train the whole model on the PIQ23 dataset.
\item Training Time: 2 hours
\item Training Strategy: Pair-wise training
\item Augmentations: Randomly crop
\end{itemize}

\begin{figure*}[ht]
    \centering
  \includegraphics[width=\textwidth]{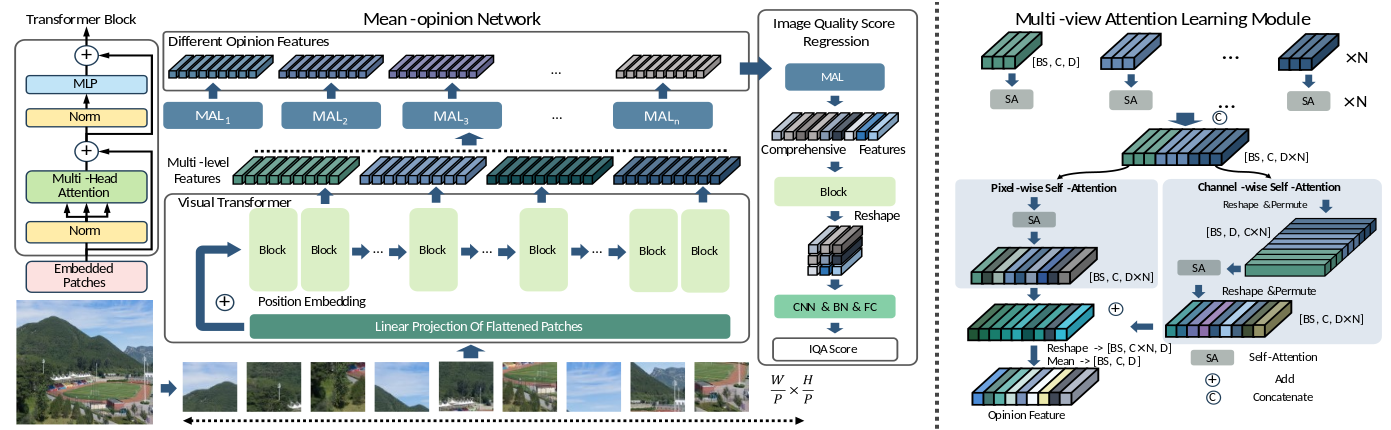}
  \caption{Network architecture of the proposed MoNet (left) and multi-view attention learning module (right).}
    \label{fig:team_1}
\end{figure*}

\subsection{A Mean-Opinion Network For Portrait Quality Assessment: MoNet}
\label{sec:monet}


\begin{center}

\vspace{2mm}
\noindent\emph{\textbf{Team I$^2$Group}}
\vspace{2mm}

\noindent\emph{Zewen Chen~$^{1,2}$,
Wen Wang~$^3$,
Juan Wang~$^1$,
Bing Li~$^1$}

\vspace{2mm}

\noindent\emph{
$^1$ State Key Laboratory of Multimodal Artificial Intelligence Systems, CASIA\\
$^2$ School of Artificial Intelligence, University of Chinese Academy of Sciences\\
$^3$ Beijing Jiaotong University}

\vspace{2mm}

\noindent{\emph{Contact: \url{chenzewen2022@ia.ac.cn}}}

\end{center}

We take the dataset annotation process, where different annotators will annotate different opinion scores for the same image and the average of theses scores is applied as the label, namely mean opinion score (MOS). Thus, a novel network architecture called mean-opinion network (MoNet) is proposed~\cite{chen2024gmc}. Mimicking the human rating process, we develop a multi-view attention learning (MAL) module for the MoNet to implicitly learn diverse opinion features by capturing complementary contexts from various perspectives. The opinion features collected from different MALs are integrated into a comprehensive quality score, effectively relieving the impacts of hyper-parameter configurations on the performance, facilitating more robust quality score assessment. To be more alignment with this challenge, we additionally take a full connection (FC) layer to get the scenes classification.

\paragraph{Global Method Description}

We present a novel network called mean-opinion network (MoNet), which collects various opinions by capturing diverse attention contexts to make a comprehensive decision on the image quality score. Fig.
\ref{fig:team_1} shows the network architecture of the MoNet, which mainly consists of three parts: i) a pre-trained ViT is employed for multi-level feature perception, ii) multi-view attention learning (MAL) modules are proposed for opinion collection, and iii) an image quality score regression module is designed for quality estimation. 

\textbf{A) Multi-level Semantic Perception.}
Given an image $I \in \mathbb{R}^{H \times W \times 3}$, we firstly crop it into $C$ patches with the size of $S \times S$, where $H$ and $W$ denote the height and width of the image and $C = \frac{H\times W}{S^2}$.
Then the patches are flattened and fed into a linear projection with the dimension of $D$, producing the embedding feature $\mathbf{E} \in \mathbb{R}^{C\times D}$. Subsequently, the features $\mathbf{E}$ sequentially traverses 12 transformer blocks, resulting in a set of multi-level features. Finally, the outputs of $N$ transformer blocks are selected and used as basic features, denoted as $f_j$ ($1\leq j \leq N$).

\textbf{B) Multi-view Attention Learning Module.}
The critical part of the MoNet is the multi-view attention learning (MAL) module. The motivation behind it is that individuals often have diverse subjective perceptions and regions of interest when viewing the same image. To this end, we employ multiple MALs to learn attentions from different viewpoints.
Each MAL is initialized with different weights and updated independently to encourage diversity and avoid redundant output features. The number of MALs can be flexibly set as a hyper-parameter. We show in our results its effect on the performance of our model.

As shown in \cref{fig:team_1}, the MAL starts from $N$ self-attentions (SAs), each of which is responsible to process a basic feature $\mathbf{f}_j$ ($1\leq j \leq N$). 
The outputs of all the SAs are concatenated, forming a multi-level aggregated feature $\mathbf{F}\in \mathbb{R}^{C\times D \times N}$. Then $\mathbf{F}$ passes through two branches, \emph{i.e.}, a pixel-wise SA branch and a channel-wise SA branch, which apply a SA across spatial and channel dimensions, respectively, to capture complementary non-local contexts and generate multi-view attention maps.
In particular, for the channel-wise SA, the feature $\mathbf{F}$ is first reshaped and permuted to convert the size from $C\times D \times N$ to $D\times (C \times N)$. After the SA, the output feature is permuted and reshaped back to the original size $C\times D \times N$. Subsequently, the outputs of the two branches are added and average pooled, generating an opinion feature.
The design of the two branches has two key advantages. 
First, implementing the SA in different dimensions promotes diverse attention learning, yielding complementary information. Second, contextualized long-range relationships are aggregated, benefiting global quality perception.

 \textbf{C) Image Quality Score Regression.}
Assuming that $M$ opinion features are generated from $M$ MALs employed in the MoNet. To derive a global quality score from the collected opinion features, we utilize an additional MAL. The MAL integrates diverse contextual perspectives, resulting in a comprehensive opinion feature that captures essential information. This feature is then processed through a transformer block, three convolutional layers with kernel sizes of $5 \times 5$, $3 \times 3$, and $3 \times 3$ to reduce the number of channels, followed by two fully connected layers that transform the feature size from 128 to 64 and from 64 to 1. Finally, we obtain a predicted quality score from the MoNet. For the scene classification, we additionally take a FC layer simiar to the image quality score regression.

\paragraph{Implementation details}

The pre-trained ViT model \texttt{vit\_base\_patch16\_384} is used as the backbone of the MoNet. We use $N=4$ transformer blocks to extract basic features, namely the 3rd, 6th, 9th and 12th blocks.
The default number of the MAL is set to $M=5$.
We use the Adam optimizer with a learning rate of $1 \times 10^{-5}$ and a weight decay of $1 \times 10^{-5}$. 
The learning rate is adjusted using the Cosine Annealing for every 50 epochs.
We train our model for 100 epochs with a batch size of 11 on one RTX3090. We take the mean square error (MSE) loss to reduce the discrepancy between the predicted scores and ground truths (GT). And we take the cross-entropy loss for scene classification.

\begin{figure*}[t]
    \centering
    \includegraphics[width=0.81\linewidth]{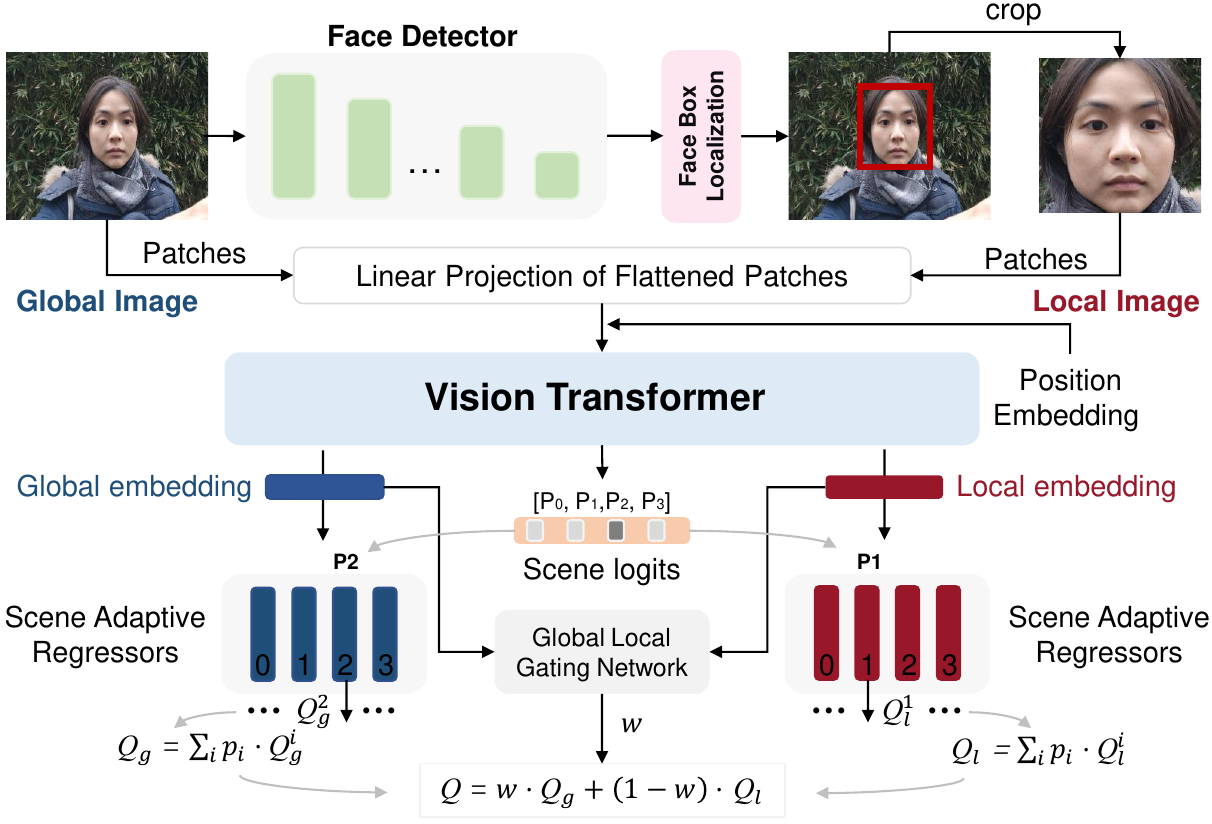} 
    \caption{Overview the method proposed by Team SECE-SYSU. The model leverages a face detector for facial localization. A Vision Transformer then extracts quality-aware embeddings from both the local facial region and the global image. Scene classification guides the selection of scene-specific regressors for global and local quality prediction. A gating network dynamically fuses these predictions for the final quality score.}
    \label{fig:team5}
\end{figure*}

\subsection{Scene Adaptive Global Context and Local Facial Perception Network for Portrait Image Quality Assessment}
\label{sec:sece}


\begin{center}

\vspace{2mm}
\noindent\emph{\textbf{Team SECE-SYSU}}
\vspace{2mm}

\noindent\emph{Xiaoqi Wang, Junqi Liu, Zixi Guo, and Yun Zhang}

\vspace{2mm}

\noindent\emph{
School of Electronics and Communication Engineering, Sun Yat-sen University, China}

\vspace{2mm}

\noindent{\emph{Contact: \url{zhangyun2@mail.sysu.edu.cn}}}

\end{center}

The facial region is pivotal for portrait image quality evaluation, yet it typically occupies only a small portion of the entire image, which poses a challenge for deep neural networks that tend to capture global semantics and context. Furthermore, scene-dependent variations in portrait quality scores introduce additional complexities~\cite{chahine2024generalized}. To address these issues, this solution proposes a scene-adaptive global context and local facial perception network. The proposed method first leverages a face detector~\cite{li2018dsfd} to precisely localize the facial region within the global image. Then, vision Transformer is employed to model quality-centric embeddings of both local facial region and global image. To address scene-specific quality biases, we formulate a scene recognition task and leverage scene category to adaptively select scene-specific global and local facial regressors. Finally, a global local gating network dynamically adjusts the weighting of quality predictions from the two branches, resulting in the final quality score.

\paragraph{Global Method Description}

We propose a novel solution that scene-adaptively evaluates the global image and detected local facial image through a face detector, ultimately fusing the local and global assessments to obtain a final quality score through a global local gating network. The proposed model, illustrated in Figure~\ref{fig:team5}, is structured as follows:
\noindent\textbf{Face Detector} employs a lightweight Dual Shot Face Detector~\cite{li2018dsfd} for robust facial localization within portrait images. The initial confidence threshold is empirically set to 0.8, achieving a 99\% face detection rate on PIQ23 database~\cite{Chahine_2023_CVPR}. In the absence of a detected face at this threshold, the Detector iteratively decreases the confidence threshold (0.7, 0.6, 0.5, 0.4, 0.2) until at least one face is successfully identified. The longer edge of the detected face bounding box determines the cropping dimension, ensuring a minimum resolution of 512x512 pixels. \noindent\textbf{Feature Extraction Module} leverages the first 10 Transformer layers from the ViT-Base~\cite{vit}, initialized with pre-trained weights from the CLIP visual encoder~\cite{clip}. The initial 6 blocks are frozen, and the remaining parameters are fine-tuned on PIQ23 database. The outputs of the first 6 blocks are fed into a convolutional layer and linear layer for scene classification. The global and local embeddings are derived by mapping the concatenated features from the last three ViT blocks to the ViT's embedding space via a convolutional projection. \noindent\textbf{Global Local Gating Network} is constructed to weight local and global quality predictions, comprising an input layer, two fully-connected hidden layers of sizes 128 and 64 with ReLU activations, and a single output neuron with a sigmoid activation. \noindent\textbf{Scene Adaptive Regressors}, implemented as linear layers, are selected based on the ground truth scene categories (training phase). The model is \textit{trained} under three scenarios: global-only, local-only, and joint global-local, with respective probabilities of 0.3, 0.3, and 0.4. During the \textit{testing} phase, global and local image are jointly processed, and their scores are weighted by the predicted scene probabilities and individual scores. The gated fusion network then weights the local and global scores to yield the final prediction. 

\vspace{-5mm}

\paragraph{Implementation details}

The proposed model was constructed using the PyTorch framework and trained on an NVIDIA GeForce RTX 3090 GPU~(24G). The experiment employed an 80-epoch training regimen with a batch size of 16. The AdamW optimizer with betas of 0.9 and 0.999 was utilized for training, initialized with a learning rate of 1e-5 and an L2 weight decay of 1e-5. A cosine annealing learning rate scheduler was adopted, with a warm-up phase reaching a maximum learning rate of 1e-4 and a minimum learning rate of 0 over 30 cycles. The objective function was the Huber loss, with a hyperparameter of 0.2. Data from the PIQ23 dataset was split, with 90\% of samples from each scene used for training and the remaining 10\% used for testing. During preprocessing, input images were subdivided into 224 $\times224$ patches. For training, a single patch was randomly sampled per image and underwent random flipping for data augmentation. The training phase took approximately 8 hours. In the testing phase, 30 patches were densely sampled from each image, and the final prediction was obtained by averaging the predicted results. The proposed model achieves an SROCC of 0.8335 and a PLCC of 0.8422 on PIQ23 following the scene-based data partitioning of Chahine et al~\cite{Chahine_2023_CVPR}. Our model achieves an inference time of 755ms per image on an Intel i5-12400F CPU with 16GB RAM and an NVIDIA GeForce RTX 2080 8GB GPU (model complexity details in Table~\ref{tab:sece_efficiency}).

\begin{table}[thbp]
	\centering
	\renewcommand{\arraystretch}{1.1}
	\resizebox{3. in}{!} 	{
		\begin{tabular}{cccc}
			\toprule
			\multicolumn{2}{c}{Module} &  Face detector & Other modules \\
                \midrule
			\multicolumn{2}{c}{MACs} & 3.36G &26.11G \\  
			\multicolumn{2}{c}{FLOPs} & 3.36G & 35.43G \\  
			\multicolumn{2}{c}{Params} & 14.22M & 135.02M \\
                \midrule
			\multicolumn{3}{l}{Inference time~(per image): 755 ms} \\   
			\bottomrule
		\end{tabular}	
	}
        \caption{Efficiency study of Team SECE-SYSU.}
	\label{tab:sece_efficiency}
\end{table}




\section*{Acknowledgements}
This work was partially supported by the Humboldt Foundation. We thank the NTIRE 2024 sponsors: Meta Reality Labs, OPPO, KuaiShou, Huawei and University of W\"urzburg (Computer Vision Lab).

The organizers thank the DXOMARK engineers and photographs for their time investment and fruitful discussions about image quality.

{\small
\bibliographystyle{ieeenat_fullname}
\bibliography{refs}
}

\end{document}